\documentclass{article} 
\usepackage{iclr2027_conference,times}

\usepackage{amsmath,amsfonts,bm}

\def\eqref#1{equation~\ref{#1}}

\def\1{\bm{1}}

\DeclareMathAlphabet{\mathsfit}{\encodingdefault}{\sfdefault}{m}{sl}
\SetMathAlphabet{\mathsfit}{bold}{\encodingdefault}{\sfdefault}{bx}{n}

\usepackage{hyperref}
\usepackage{url}
\usepackage{graphicx}
\usepackage{amsmath}
\usepackage{amssymb}

\usepackage{url}

\usepackage{booktabs}
\usepackage{microtype}
\usepackage{graphicx}
\usepackage{booktabs} 
\usepackage{makecell}
\usepackage{amsmath}

\usepackage{verbatim}
\usepackage{epstopdf}
\usepackage{multirow}
\usepackage{amsfonts}
\usepackage{soul}
\usepackage{graphicx}
\usepackage{subcaption} 
\usepackage{newfloat}
\usepackage{listings}
\usepackage[table]{xcolor}
\usepackage{wrapfig}
\usepackage[most]{tcolorbox}

\title{
Fast and Accurate: Improving Personalized LLM Performance by Semantic Concentration Inspired Workload Split between RAG and Fine-Tuning
}

\author{
\textbf{Xingke Yang$^{1}$,
Sicong Li$^{1}$,
Lingfeng Yao$^{1}$,
Rui Chen$^{2}$,
Changqing Luo$^{1}$,
}
\\
\textbf{
Xin Fu$^{1}$,
Miao Pan$^{1}$
}
\\[0.5em]
$^{1}$University of Houston, USA \quad
$^{2}$Independent Researcher, USA \quad
}

\iclrfinalcopy 
\begin{document}

\maketitle
\lhead{Preprint}

\begin{abstract}
\textcolor{black}{With ever increasing personalized large language model (LLM) services, the next question to answer is how to improve the quality of services/users' experiences, which is indicated by personalized LLM generation accuracy/quality and inference latency. Although state-of-the-art designs, e.g., parameter-efficient fine-tuning (PEFT), retrieval-augmented generation (RAG) or Full PEFT+RAG, can help, they largely ignore the heterogeneous semantic concentrations of users' historical data and cannot fully unleash the potential of personal data for personalized LLM performance enhancement. In this paper, we propose PEFT Union RAG (PUR), a novel ``fast and accurate'' approach that exploits the semantic concentration of users' personal historical data to split workload between PEFT and RAG, aiming to improve personalized LLM performance in terms of better generation accuracy/quality and lower inference latency. Briefly, PUR partitions user's personal historical data and assigns semantic concentrated ``core'' information to PEFT for offline parametric fine-tuning, while retaining dispersed information for RAG. During the online LLM inference, PUR directly generates responses for ``core'' queries covered by the PEFT region and invokes RAG only for sporadic ``peripheral'' queries. Experimental results show that, by appropriately splitting workload between PEFT and RAG, PUR consistently achieves the best generation accuracy/quality for various users across multiple tasks, while reducing inference latency by 14.4\%--46.6\% compared with the best performed method.}

\end{abstract}

\section{Introduction}


\textcolor{black}{While large language models (LLMs) are increasingly employed for user-facing applications such as email drafting~\citep{miura2025understanding}, recommendation~\citep{zhu2024collaborative}, and content generation~\citep{tan2024proxyqa}, the general responses from base LLMs may not be good enough to meet users’ personalized expectations. Nowadays, the concept of personalized LLM is becoming more and more popular, which leverages user's personal historical data to enhance LLMs, so that the generated responses can better align with users’ specific needs and preferences~\citep{tan2024personalized,liu2025llms+}. Under such problem settings, the essential question to ask is how to effectively exploit the user's historical data to improve personalized LLM's performance.}

\textcolor{black}{To enhance personalized LLM's accuracy,}
\textcolor{black}{existing approaches can be broadly divided into two categories~\citep{salemi2025comparing,tan2024democratizing}: retrieval-augmented generation (RAG) based methods and parameter-efficient fine-tuning (PEFT) based methods. The former treats user historical data as a reference corpus and retrieves content relevant to the current user prompt during inference, using the retrieved content as additional context to guide generation. The latter fine-tunes a small set of trainable parameters based on user historical data, allowing the model to encode user-specific preferences or behavioral patterns into lightweight personalized parameters that can support personalized generation during inference.}

\begin{figure}[t]
    \centering

    \begin{minipage}[t]{0.66\linewidth}
        \centering
        \begin{subfigure}[t]{0.48\linewidth}
            \centering
            \includegraphics[width=\linewidth]{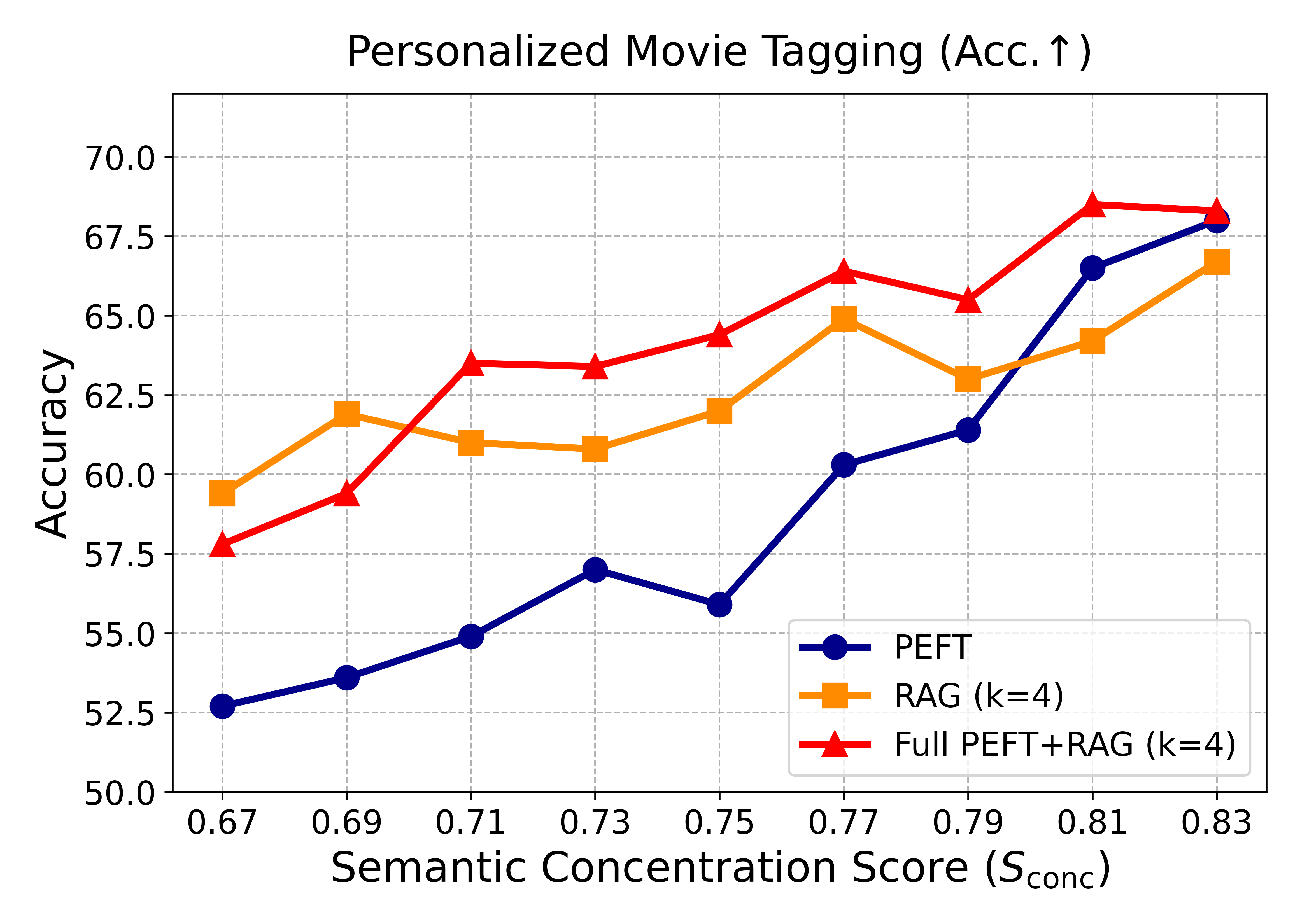}
        \end{subfigure}
        \hfill
        \begin{subfigure}[t]{0.48\linewidth}
            \centering
            \includegraphics[width=\linewidth]{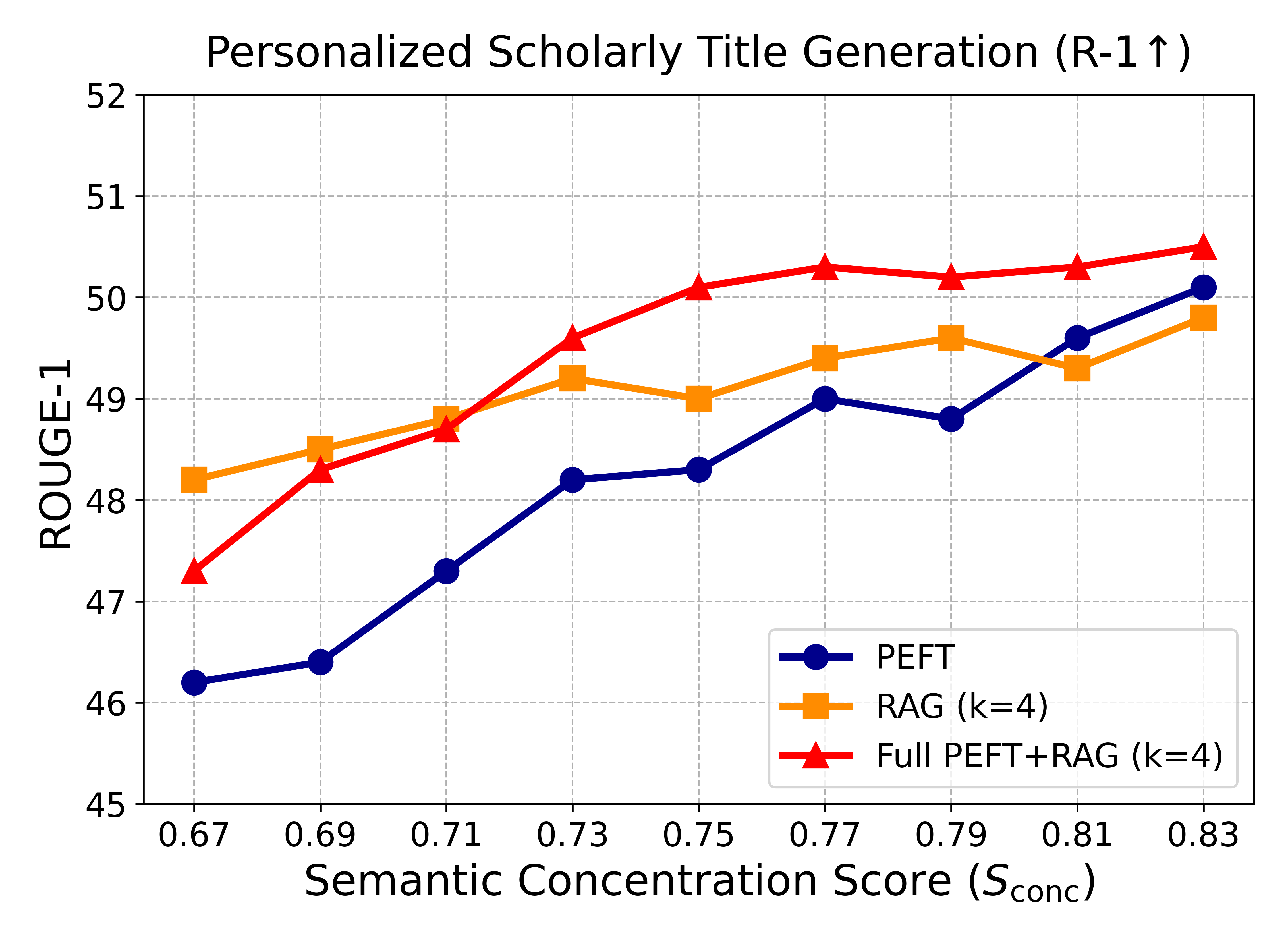}
        \end{subfigure}

        \caption*{(a) Generation accuracy/quality.}
    \end{minipage}
    \hfill
    \begin{subfigure}[t]{0.32\linewidth}
        \centering
        \includegraphics[width=\linewidth]{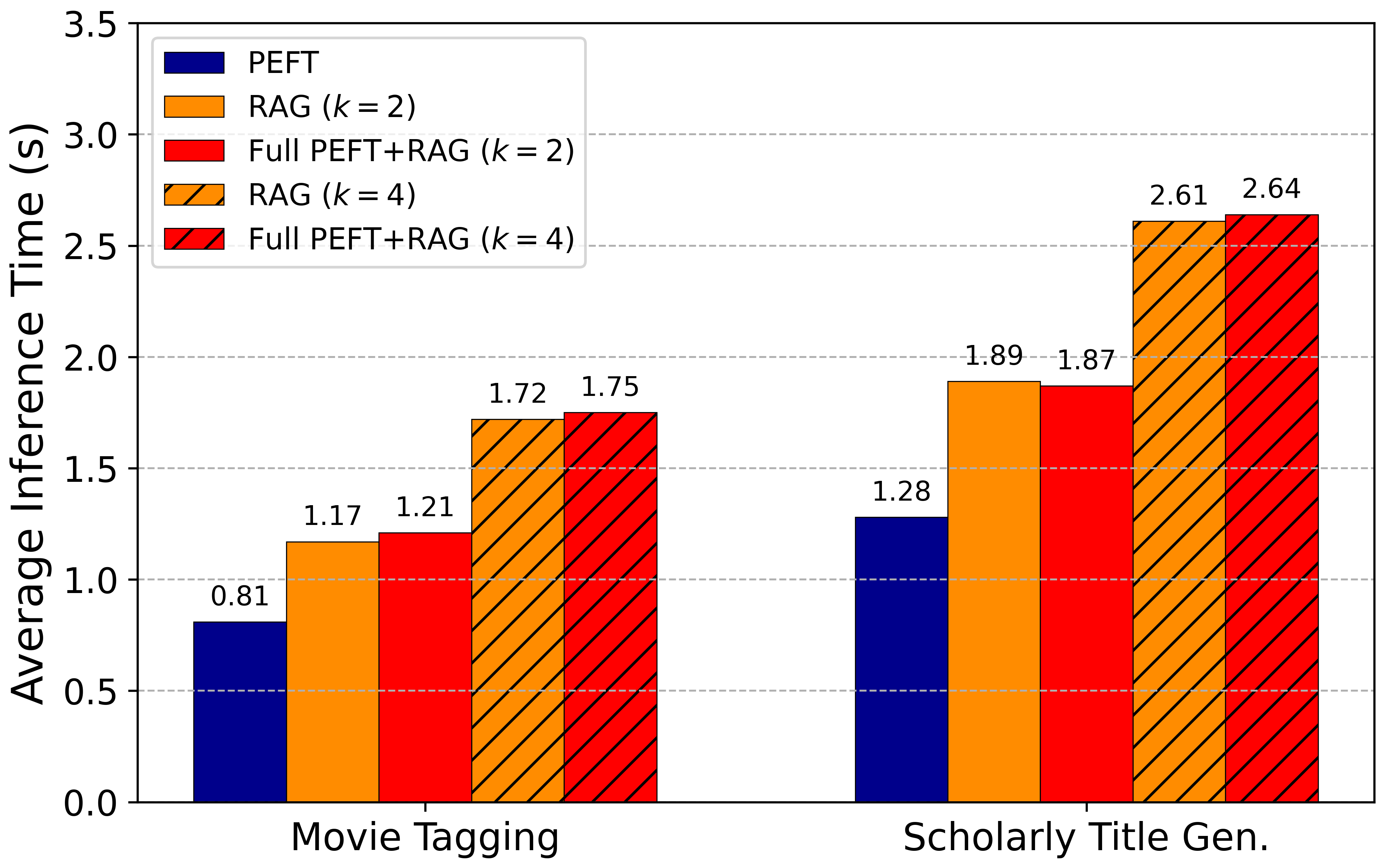}
        \caption*{(b) Inference latency.}
        \label{fig:motivation_latency}
    \end{subfigure}

    \caption{\textcolor{black}{Preliminary results. (a) The generation accuracy/quality of PEFT, RAG, and Full PEFT+RAG across users with different semantic concentration scores. (b) Average inference latency under different methods, where k denotes the number of retrieved texts used at inference.}}
    \label{fig:motivation_overview}
    \vspace{-2em}
\end{figure}

\textcolor{black}{To answer the question which one (i.e., RAG or PEFT) can better enhance personalized LLM's performance, prior studies~\citep{salemi2025comparing,lakatos2025investigating,ovadia2024fine} have extensively compared these two paradigms across multiple personalized LLM tasks involving large numbers of users with personal historical data. Those works show that RAG often serves as a better option and outperforms PEFT in many settings. Motivated by the complementary strengths of retrieval/RAG and parametric adaptation/PEFT, some other studies~\citep{tan2024democratizing,ficek2024gpt,pingua2025medical} further explore to use them both fully (i.e., full PEFT+RAG = 100\% PEFT + 100\% RAG). Those methods first fully fine-tune the model with user's historical data and then retrieve all relevant user's content as additional contexts during the inference. Experimental results show that full PEFT+RAG can further improve personalized LLM's performance.}

\textcolor{black}{However, the state-of-the-art research findings above are based on the averaged performance results over all users, which largely ignores the heterogeneity of users' historical data. Actually, for personalized LLM tasks, users’ historical data may substantially differ in semantic concentration. Some users may have more concentrated historical data, reflecting stronger consistency and stability, whereas some other users may have more dispersed historical data that exhibit greater sparsity and stronger long-tail characteristics. As shown in Fig.~\ref{fig:motivation_overview} (a), our empirical studies indicate that prior findings are not applicable, if users are grouped by the semantic concentration levels of their historical data. Specifically, for users with very low semantic concentration, full PEFT+RAG is inferior to RAG alone, while for users with high semantic concentration, PEFT no longer lags behind RAG. Those results motivate us to thoroughly consider the semantic concentration of users' historical data and seek a solution to better integrate PEFT and RAG, to further improve personalized LLM performance.}

\textcolor{black}{Besides, as the personalized LLM is being deployed for daily services, the LLM performance should consider not only LLM generation accuracy/quality but also online LLM inference latency for good quality of services. Due to the extra historical corpus retrieval step and the additional computation introduced by long contexts, either RAG or full PEFT+RAG incurs much higher inference latency than directly generating responses from the prompt~\citep{jiang2025rago,shen2025hermes,mahapatra2025storage}. As shown in Fig.~\ref{fig:motivation_overview} (b), for methods involving RAG, inference latency significantly increases as more reference texts are retrieved and incorporated into the input context. It is obvious that the benefits brought by RAG come at the cost of severely delayed LLM response.}

\textcolor{black}{To address the observed challenges above, in this paper, we propose PEFT Union RAG (PUR), a novel ``fast and accurate'' approach that exploits the semantic concentration of users' personal historical data to split workload between PEFT and RAG, aiming to improve personalized LLM performance in terms of better generation accuracy/quality and lower inference latency. Unlike full PEFT+RAG, which directly combines 100\% PEFT and 100\% RAG, PUR first quantifies the semantic concentration of the user’s historical data and then partitions the user’s historical data accordingly. Then, PUR allocates semantically concentrated core information to offline PEFT for parametric adaptation, and saves dispersed information as an external corpus for RAG.} \textcolor{black}{At the online LLM inference stage, the personalized LLM checks if the knowledge required by the current user query has already been captured by the fine-tuned parameters. If so, the personalized LLM will directly generate the response. Otherwise, the personalized LLM will follow the traditional RAG protocol to generate the response. A sketch of the proposed PUR design is presented in Fig.~\ref{fig:overview}.}

\textcolor{black}{Such semantic concentration based workload split in PUR makes the best use of PEFT's and RAG's complementary advantages. Briefly, by restricting PEFT training to highly concentrated core historical data, PUR reduces the noise introduced by dispersed information and thereby improves response accuracy/quality for frequent ``core'' queries. For those sporadic ``peripheral'' queries, PUR still employs RAG to provide additional support to keep personalized LLM response quality. Meanwhile, by allowing ``core'' queries to bypass unnecessary RAG, PUR helps to effectively reduce the averaged online inference latency. Note that the time consumed by PEFT is offline fine-tuning cost, and should not be counted as part of online LLM inference delay that affects quality of experience for users. Extensive experimental results show that PUR yields the best generation accuracy/quality for users with different semantic concentration levels across various tasks, while reducing inference latency by 14.4\%--46.6\% compared with the best performed compared method.}

\section{Preliminaries}
\textcolor{black}{The goal of personalized LLM generation is to leverage user historical data to generate outputs that better align with the user's preferences or behavioral patterns. Formally, let $M_{\theta}$ denote an LLM with parameters $\theta$. For a user $u$, we denote the historical data by $H_u=\{h_1,h_2,\ldots,h_{N_u}\}$, where $N_u$ is the number of historical data instances. Given the current user input $x_u$, personalized generation can be written as $y_u = M_{\theta}(x_u \mid H_u)$, where the output $y_u$ depends on both the current input and the user's history. Under this formulation, the key challenge is how to effectively incorporate $H_u$ into personalized generation.}

\textcolor{black}{\noindent\textbf{PEFT enhanced personalized LLM.}} 
\textcolor{black}{Parameter-Efficient Fine-Tuning (PEFT) is a class of methods for adapting LLMs to downstream tasks with low training cost, with representative approaches including Adapter~\citep{houlsby2019parameter}, LoRA~\citep{hu2022lora}, and BitFit~\citep{zaken2022bitfit}. The core idea of PEFT is to update only a small set of additional trainable parameters or lightweight modules, rather than fine-tuning all parameters.}

\textcolor{black}{In PEFT enhanced personalized LLM, the user's historical data $H_u$ is used to train user-specific PEFT parameters $\phi_u$ during the offline stage. Let $M_\theta$ denote the base model with frozen pretrained parameters $\theta$. The user-specific adaptation objective can then be written as
$
\phi_u^*=\arg\min_{\phi_u}\mathcal{L}\big(M_{\theta,\phi_u};H_u\big),
$
where $\mathcal{L}$ denotes the task-specific loss on $H_u$, and $M_{\theta,\phi_u}$ is the personalized model obtained by incorporating $\phi_u$ into the base model. During the online LLM inference, the personalized model directly maps the current input $x_u$ to the personalized output:}
$$
y_u=M_{\theta,\phi_u^*}(x_u).
$$

\textcolor{black}{\noindent\textbf{RAG enhanced personalized LLM.}} 
\textcolor{black}{
Retrieval-Augmented Generation (RAG)~\citep{lewis2020retrieval,zamani2022retrieval} is a paradigm that augments LLM generation with external retrieval. Unlike PEFT, which incorporates knowledge into model parameters, RAG stores such knowledge in an external database and retrieves relevant content during inference based on the current input. The retrieved content is then provided to the LLM as additional context to improve the relevance and factual accuracy of the generated output, making RAG particularly effective for knowledge-intensive tasks. In modern RAG systems, retrieval is typically implemented by encoding text into a vector space with a semantic embedding model and returning the most relevant items based on vector similarity~\citep{yu2024rankrag,shao2024scaling}.}

\textcolor{black}{In RAG enhanced personalized LLM, the user historical data $H_u$ is treated as a personalized knowledge source. In the offline stage, each historical record $h_i$ is encoded into a vector representation to build a user-specific retrieval database. During the inference, given the current input $x_u$, the retriever returns the top-$k$ most relevant historical records from $H_u$, denoted by $R_k(x_u, H_u)$. The base model then takes $x_u$ together with the retrieved results as context and produces the personalized output:}
\[
y_u = M_\theta(x_u, R_k(x_u, H_u)).
\]

\textcolor{black}{\noindent\textbf{100\% PEFT + 100\% RAG enhanced personalized LLM.}}
\textcolor{black}{
In addition to the two paradigms above, prior works~\citep{tan2024democratizing,ficek2024gpt,pingua2025medical} have also proposed a method that directly combines PEFT and RAG to further improve personalized generation. Specifically, in the offline stage, the full user historical data $H_u$ is used both to train user-specific PEFT parameters $\phi_u$ and to build a user-specific retrieval database. During the online LLM inference, the system first performs retrieval and then uses the personalized model $M_{\theta,\phi_u^*}$ to generate the personalized output:}
\[
y_u = M_{\theta,\phi_u^*}(x_u, R_k(x_u, H_u)).
\]

\section{Observations and Motivation}

\textcolor{black}{To examine how the above methods behave for users with different historical data distributions, we conducted preliminary experiments on the personalized movie tagging (text classification) and the personalized scholarly title generation (text generation) tasks. We grouped users according to the semantic concentration (Sec.~\ref{sec:Semantic Concentration Modeling}) of their historical data in the embedding space, and measured both generation quality and inference latency under different methods. The results are shown in Fig.~\ref{fig:motivation_overview}.}

\textcolor{black}{\noindent\textbf{No Strategy Always Wins.} As shown in Fig.~\ref{fig:motivation_overview} (a), we observe the following pattern in both tasks. The method that achieves the best task performance varies across users with different levels of semantic concentration, indicating that no single strategy serves as the universal best choice for all users. Moreover, PEFT appears to be particularly sensitive to semantic concentration. For users with lower semantic concentration, PEFT performs noticeably worse than RAG, and even the Full PEFT+RAG strategy fails to outperform RAG alone. In contrast, as semantic concentration increases, PEFT becomes increasingly competitive and no longer lags behind RAG.}

\textcolor{black}{\noindent\textbf{Retrieval Slows Down Inference.} Retrieval-augmented strategies incur additional inference latency due to the extra retrieval step and longer input contexts. In personalized tasks, because the per-user historical corpus is relatively small, the retrieval step itself is typically not the dominant cost; instead, the main overhead comes from the longer context introduced by retrieved texts. As shown in Fig.~\ref{fig:motivation_overview} (b), when Top-$k=4$, RAG-related methods increase inference latency by more than 100\% compared to the PEFT-only method. }

\textcolor{black}{Those results motivate us to thoroughly consider the semantic concentration of users' historical data and seek a solution to better integrate PEFT and RAG, to achieve both better generation accuracy/quality and low inference latency.}

\section{PEFT Union RAG Design}

\begin{figure*}
    \centering
    \includegraphics[width=0.99\textwidth]{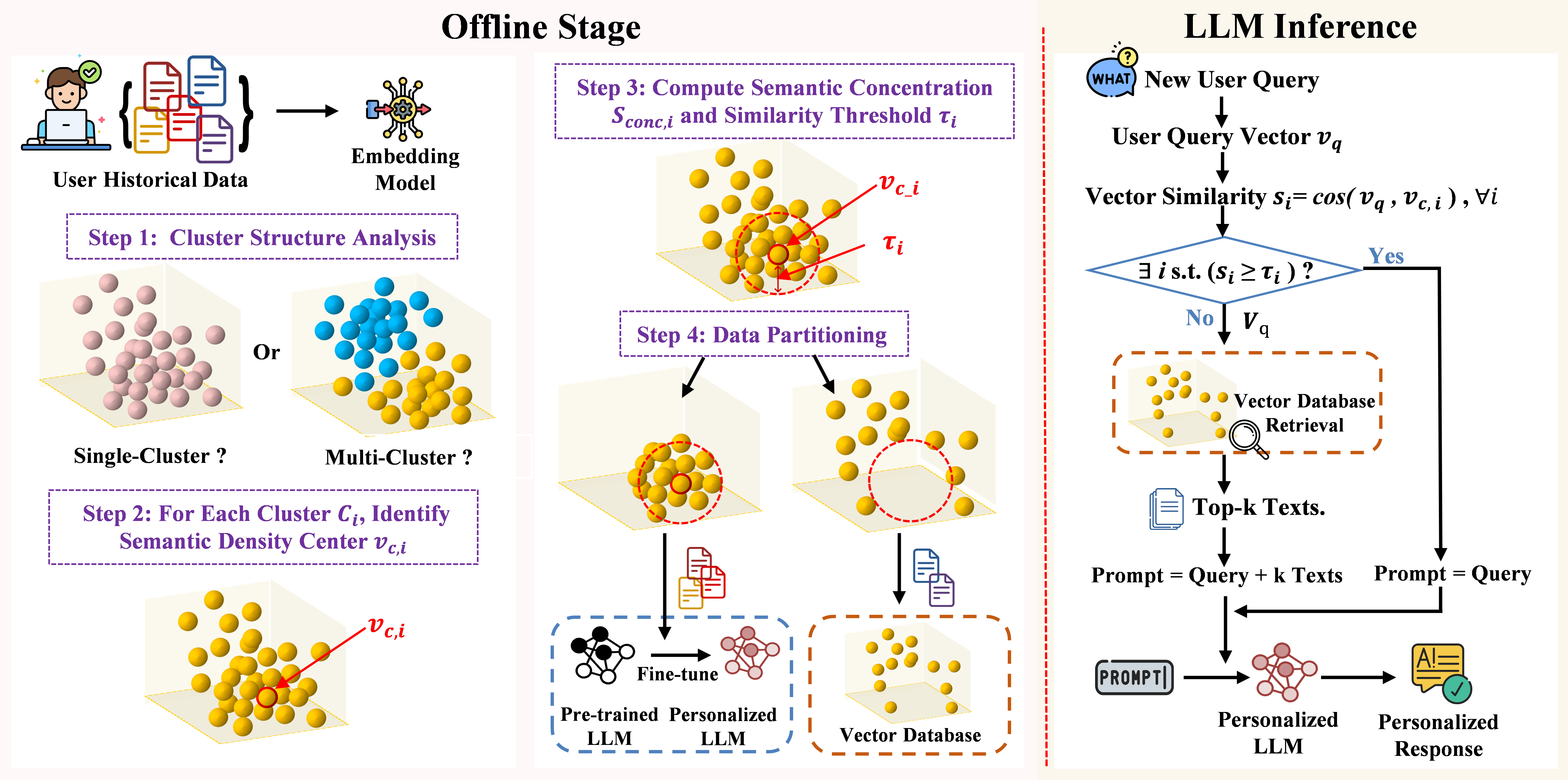}
    \caption{\textcolor{black}{An overview of our proposed PEFT Union RAG (PUR). }}
    \label{fig:overview}
    \vspace{-1.0em}
\end{figure*}

\textcolor{black}{Based on the observations above, we propose PEFT Union RAG (PUR), a novel ``fast and accurate'' approach that exploits the semantic concentration of users' personal historical data to split workload between PEFT and RAG, aiming to improve personalized LLM performance in terms of better generation accuracy/quality and lower inference latency. Figure~\ref{fig:overview} illustrates the overall workflow of PUR. At the offline stage, unlike directly applying full PEFT+RAG to the entire user's historical data, PUR first quantifies the semantic concentration of the user’s historical data and then partitions it accordingly. The more semantically concentrated ``core'' subset is allocated to PEFT for parametric learning, while the remaining dispersed subset is retained in the reference corpus for RAG. During the online LLM inference, the personalized LLM checks if the knowledge required by the current user query has already been captured by the fine-tuned parameters. If so, the personalized LLM will directly generate the response. Otherwise, the personalized LLM will follow the traditional RAG protocol to generate the response. The following subsections will illustrate the proposed PUR design in detail.}

\subsection{Semantic Concentration Modeling}\label{sec:Semantic Concentration Modeling}

\textcolor{black}{\noindent\textbf{Historical Data Embedding.} In PUR, we analyze the semantic concentration of user's historical data in the embedding space. Since local neighborhood structure in the embedding space can reflect semantic similarity between historical records, we first use an embedding model to encode the historical records into vector representations. This mirrors the embedding step commonly used in RAG systems to build a vector database. For user $u$ with historical data $H_u=\{h_1,\ldots,h_{N_u}\}$, we encode each historical record $h_i$ into a vector $v_i$, and obtain the embedding set $V_u=\{v_1,\ldots,v_{N_u}\}$.}

\textcolor{black}{\noindent\textbf{Cluster Structure Analysis.}
After obtaining the embedding set $V_u$, we use HDBSCAN~\citep{campello2015hierarchical} to identify its cluster structure. This method can detect density-based clusters without requiring the number of clusters to be specified in advance. This step is necessary because our subsequent definition of semantic concentration is based on a single semantic density center. If multiple semantic centers exist, a single global concentration score would mix within-cluster dispersion with between-cluster separation, leading to a distorted estimate under the single-center assumption. Therefore, when multiple clusters are detected, we conduct the subsequent concentration analysis separately within each cluster $\mathcal{C}_i$.}

\textcolor{black}{\noindent\textbf{Semantic Density Center Identification.} Semantic concentration aims to measure how closely samples are distributed around the core semantic region, and therefore depends on an appropriate semantic center. A natural choice is the centroid, but this choice is sensitive to tail points: samples far from the dominant region may substantially bias the centroid and pull it away from the core semantic region. Therefore, to obtain a more robust center estimate, PUR identifies the semantic density center for each cluster based on local density. Specifically, for each sample $v_j \in \mathcal{C}_i$, we compute its $k$-NN radius~\citep{cover1967nearest}, i.e., the distance to its $k$-th nearest neighbor, denoted by $r_j^{(k)}$. This metric serves as a proxy for local density, where smaller values indicate denser local neighborhoods. We then define the sample with the smallest $k$-NN radius as the semantic density center of cluster $\mathcal{C}_i$, denoted by $v_{c,i}$. This center is more likely to lie in the densest and most stable semantic region of the cluster.}

\textcolor{black}{\textbf{Semantic Concentration.} After estimating the semantic density center $v_{c,i}$, we define a cluster-level semantic concentration score $S_{\mathrm{conc},i}$. For each embedding vector $v_j \in \mathcal{C}_i$, we measure its proximity to the density center using cosine similarity, i.e., $s_j = \cos(v_j, v_{c,i})$. A larger $s_j$ indicates that the sample is semantically closer to the center. Based on these similarities, we first define the average concentration term as $S_{\mathrm{avg},i} = \frac{1}{|\mathcal{C}_i|} \sum_{v_j \in \mathcal{C}_i} s_j$, which measures the average proximity of cluster samples to the density center. A higher value indicates that the cluster is more concentrated around its semantic core.}

\textcolor{black}{However, average similarity alone may underestimate clusters that contain a compact dominant core together with a long tail. In such cases, low-similarity tail samples can substantially drag down the average, causing the $S_{\mathrm{avg},i}$ to underestimate the PEFT suitability of the dominant core. To address this issue, we introduce a tail-aware compensation term that fixes the underestimation issue caused by pronounced long-tail structure, rather than rewarding dispersed points.}


\textcolor{black}{Specifically, we use the robust z-score~\citep{iglewicz1993volume} $z_j = (\mathrm{median}(s) - s_j) / (\mathrm{MAD}(s) + \epsilon)$, where $\mathrm{MAD}(s) = \mathrm{median}\bigl(|s_j - \mathrm{median}(s)|\bigr)$ to identify samples that lie clearly outside the dominant part of the similarity distribution. If a sample has much lower similarity than most others, its robust z-score will be larger. We then get the tail ratio as
$\pi_{\mathrm{tail},i} = \Pr (z_j > 3)$, where $\Pr$ denotes the proportion of samples that meet the condition. This term measures how many samples are clearly in the low-similarity tail, and the threshold \(3\) is a commonly used standard in robust outlier detection~\citep{iglewicz1993volume}. We further define the tail strength as $h_{\mathrm{tail},i}=Q_{50}(s)-Q_5(s),$ where \(Q_p(s)\) denotes the \(p\)-th quantile of the similarity score \(s_j\). This term measures how far the lower tail is separated from the dominant region. In this way, \(\pi_{\mathrm{tail},i}\) captures how many tail samples there are, while \(h_{\mathrm{tail},i}\) captures how strong the tail is. Their product therefore naturally reflects the overall influence of the long-tail structure. The final semantic concentration score is then defined as}
\[
S_{\mathrm{conc},i}
=
S_{\mathrm{avg},i}
+\lambda\,\pi_{\mathrm{tail},i}\,h_{\mathrm{tail},i},
\]
\textcolor{black}{where $\lambda$ controls the strength of the compensation. In this way, the score better preserves the contribution of a compact dominant core in long-tailed clusters, thereby yielding a subsequent data partitioning between PEFT and RAG that is more consistent with the underlying semantic structure.}

\subsection{Semantic Concentration Inspired Data Partitioning and Workload Split in PUR}

\textcolor{black}{After obtaining the concentration score $S_{\mathrm{conc},i}$ for each semantic cluster, PUR determines the PEFT allocation ratio for that cluster based on this score. Intuitively, a higher $S_{\mathrm{conc},i}$ indicates that a larger proportion of samples in the cluster is suitable for PEFT-based parametric learning. However, there is no closed-form mapping from $S_{\mathrm{conc},i}$ to the optimal allocation ratio. Instead, we adopt an empirically calibrated allocation strategy. Specifically, PUR evaluates candidate allocation ratios in several semantic concentration intervals and summarizes appropriate ratios in a lookup table, as shown in Table~\ref{tab:lookup_peft_ratio}. PUR does not attempt to provide an exact optimal allocation for every $S_{\mathrm{conc},i}$, but provides a robust and practical allocation policy to guide the data allocation. As shown in the subsequent experiments, the calibrated ratios generalize well across multiple tasks and diverse users.}

\textcolor{black}{After obtaining the allocation ratio $\rho_i$ for the cluster $\mathcal{C}_i$, PUR employs PEFT to fine-tune the top fraction $\rho_i$ of samples ranked by similarity to the density center. The minimum similarity value among the selected samples is defined as the cluster-level similarity threshold $\tau_i$, which is used later for LLM inference path selection. The remaining data samples are retained in the retrieval corpus to support RAG. After processing all clusters, the user's historical data is partitioned into two disjoint subsets, $H_u^{\mathrm{peft}}$ and $H_u^{\mathrm{rag}}$, where $H_u = H_u^{\mathrm{peft}} \cup H_u^{\mathrm{rag}}$. Here, $H_u^{\mathrm{peft}}$ is used to fine-tune personalized LLM parameters $\phi_u$, while $H_u^{\mathrm{rag}}$ is used to build the personalized retrieval vector database.}

\subsection{
Personalized LLM Inference in PUR}

\textcolor{black}{At the online LLM inference stage, PUR will choose the ``fast and accurate'' inference path given a user's new query. For a query $x_u$, the PUR system first encodes it into a query vector $v_q$ and computes its cosine similarity to each semantic density center $v_{c,i}$: $s_i=\cos(v_q,v_{c,i}), \quad \forall i$. If there exists a cluster such that $s_i \ge \tau_i$, the input is considered to fall within a PEFT-covered core semantic region, and the personalized model $M_{\theta,\phi_u^*}$ directly generates the response without retrieval. Otherwise, the PUR system retrieves relevant historical information from the external corpus $H_u^{\mathrm{rag}}$ and performs retrieval-augmented generation. Formally,}
\[
y_u=
\begin{cases}
M_{\theta,\phi_u^*}(x_u), & \exists i \text{ such that } s_i\ge \tau_i,\\[4pt]
M_{\theta,\phi_u^*}(x_u, R_k(x_u,H_u^{\mathrm{rag}})), & \text{otherwise.}
\end{cases}
\]


\section{Experimental Setup}

\textcolor{black}{\noindent\textbf{Datasets.} We use data from the Large Language Model Personalization (LaMP) benchmark~\citep{salemi2024lamp}, which includes seven personalized LLM tasks. Each user in the benchmark comes with its own historical data, collected from real-world user data.\footnote{We exclude LaMP-6 as it involves private data that we cannot access. Dataset statistics are presented in Appendix~\ref{sec:Dataset Details and Hyperparameters}.}  We report accuracy for classification tasks (citation identification and movie tagging), RMSE for product rating task, and ROUGE-1 for text generation tasks (news headline, scholarly title, tweet paraphrasing). Note that all metrics are the higher the better, except for RMSE.}

\textcolor{black}{\noindent\textbf{Methods for comparison.} We compare PUR with three methods for personalized generation: (1) Retrieval Augmented Generation \textbf{(RAG)}, which uses \texttt{bge-base-en-v1.5}~\citep{xiao2024c} as the embedding model and retrieves the Top $k=4$ texts by default; (2) Parameter-Efficient Fine-Tuning \textbf{(PEFT)}, implemented using LoRA~\citep{hu2022lora} with default rank $r=16$, applied to the $q\_proj$ and $v\_proj$ modules, with learning rate $1\times10^{-5}$, batch size 4, and 3 training epochs; (3) and the 100\% PEFT + 100\% RAG \textbf{(Full PEFT+RAG)}. For PUR and all methods, we adopt \texttt{Llama-2-7B}~\citep{touvron2023llama} as the base LLM.} Note that PUR focuses on workload allocation between parametric adaptation and retrieval augmentation, for which we adopt standard PEFT and RAG implementations. More advanced adaptation or retrieval systems~\citep{li2023towards,jeong2024adaptive} are orthogonal to this allocation mechanism and can be incorporated into PUR.

\textcolor{black}{\noindent\textbf{Implementation.} All experiments are conducted on a server equipped with 4 NVIDIA RTX8000 GPUs (48 GB each). For PUR, we use \texttt{bge-base-en-v1.5} as the embedding model. Under our HDBSCAN setting with \texttt{min\_cluster\_size} $= 0.3N_u$, the historical data of a single LaMP user is typically identified as a single cluster. Therefore, the main experiments use one $S_{\mathrm{conc}}$ score per user. For completeness, we further analyze multi-cluster settings using combined user data in Appendix~\ref{sec:Applicability to Multi-Cluster Historical Data}. We set the KNN parameter $k$ to 5 and $\lambda=1$. Across all users in the benchmark, the semantic concentration score $S_{\mathrm{conc}}$ ranges from 0.66 to 0.84. For grouped analysis, we partition users into intervals of width 0.02 and select 10 users with the largest historical data size from each interval to ensure more stable evaluation. For task evaluation, we use 20\% of each user's data as evaluation data and treat the remaining 80\% as historical data $H_u$.}

\begin{table*}[t]
\centering
\caption{\textcolor{black}{Performance under different semantic concentration range and PEFT allocation ratios. For each user group, we compare other methods with PUR instantiated using PEFT allocation ratios from 10\% to 90\%, to examine how the optimal allocation varies with user semantic concentration. The best score is in \textbf{bold} and the second best is \underline{underlined}.}}
\label{tab:sconc_allocation}
\resizebox{\textwidth}{!}{
\begin{tabular}{lccc|ccccccccc}
\toprule
\multirow{2}{*}{\makecell{$S_{\mathrm{conc}}$ \\ Range}}
& \multicolumn{3}{c|}{Compared Methods}
& \multicolumn{9}{c}{PUR (PEFT Allocated Ratio)} \\
\cmidrule(lr){2-4} \cmidrule(lr){5-13}
& PEFT & RAG & Full PEFT+RAG
& 10\% & 20\% & 30\% & 40\% & 50\% & 60\% & 70\% & 80\% & 90\% \\
\midrule
\multicolumn{13}{c}{\textbf{Personalized Movie Tagging (Acc. $\uparrow$)}} \\
\midrule
0.66--0.68 & 52.7 & 59.4 & 57.8 & 59.4 & 59.7 & \textbf{60.9} & \underline{59.8} & 58.1 & 56.4 & 56.0 & 54.9 & 53.3 \\
0.68--0.70 & 53.6 & 61.9 & 59.4 & 62.0 & \underline{63.9} & \textbf{64.6} & 62.5 & 60.7 & 57.9 & 57.1 & 55.2 & 53.1 \\
0.70--0.72 & 54.9 & 61.0 & 63.5 & 60.8 & 62.5 & 63.0 & \textbf{64.4} & \underline{63.7} & 61.8 & 59.6 & 57.1 & 55.8 \\
0.72--0.74 & 57.0 & 60.8 & 63.4 & 61.4 & 61.9 & 63.3 & \underline{64.1} & \textbf{65.4} & 62.2 & 60.0 & 58.9 & 57.0 \\
0.74--0.76 & 55.9 & 62.0 & 64.4 & 62.7 & 62.7 & 63.8 & 64.3 & \underline{65.3} & \textbf{66.9} & 64.4 & 60.2 & 58.1 \\
0.76--0.78 & 60.3 & 64.9 & 66.4 & 64.7 & 65.4 & 65.7 & 64.8 & 67.2 & \textbf{69.0} & \underline{67.9} & 66.4 & 62.8 \\
0.78--0.80 & 61.4 & 63.0 & 65.5 & 63.6 & 64.0 & 64.5 & 66.4 & 65.5 & \underline{68.3} & \textbf{69.1} & 67.3 & 65.2 \\
0.80--0.82 & 66.5 & 64.2 & 68.5 & 65.7 & 66.9 & 66.7 & 67.0 & 67.6 & 68.8 & 68.5 & \textbf{70.4} & \underline{69.9} \\
0.82--0.84 & 68.0 & 66.7 & 68.3 & 66.3 & 67.1 & 67.7 & 68.0 & 68.3 & 68.4 & \underline{68.9} & \textbf{69.2} & 68.4 \\
\midrule
\multicolumn{13}{c}{\textbf{Personalized Scholarly Title Generation (R-1 $\uparrow$)}} \\
\midrule
0.66--0.68 & 46.2 & 48.2 & 47.3 & 47.1 & \underline{48.7} & \textbf{49.1} & \underline{48.7} & 48.4 & 48.1 & 47.8 & 46.8 & 46.4 \\
0.68--0.70 & 46.4 & 48.5 & 48.3 & 48.3 & 48.6 & \underline{49.2} & \textbf{49.4} & 48.7 & 48.3 & 48.1 & 47.5 & 47.1 \\
0.70--0.72 & 47.3 & 48.8 & 48.7 & 48.6 & 49.2 & \underline{49.7} & \textbf{50.3} & 49.4 & 49.1 & 48.9 & 48.3 & 47.9 \\
0.72--0.74 & 48.2 & 49.2 & 49.6 & 48.8 & 48.8 & 48.9 & 49.3 & \textbf{51.2} & \underline{50.7} & 49.8 & 48.8 & 48.7 \\
0.74--0.76 & 48.3 & 49.0 & 50.1 & 49.1 & 49.0 & 49.3 & 49.8 & 50.3 & \textbf{51.5} & \underline{51.2} & 49.8 & 49.0 \\
0.76--0.78 & 49.0 & 49.4 & 50.3 & 49.4 & 49.0 & 49.4 & 49.7 & 51.3 & \underline{51.5} & \textbf{51.6} & 51.3 & 50.2 \\
0.78--0.80 & 48.8 & 49.6 & 50.2 & 48.7 & 49.9 & 49.8 & 50.9 & 50.5 & 51.2 & \textbf{51.9} & \underline{51.5} & 51.0 \\
0.80--0.82 & 49.6 & 49.3 & 50.3 & 49.3 & 49.7 & 49.9 & 50.3 & 50.2 & 50.9 & \underline{51.2} & \textbf{51.7} & 50.9 \\
0.82--0.84 & 50.1 & 49.8 & 50.5 & 49.7 & 49.9 & 49.9 & 50.0 & 50.4 & 50.4 & \underline{50.7} & \textbf{51.2} & 50.5 \\
\bottomrule
\end{tabular}
}
\vspace{-1.0em}
\end{table*}

\section{Evaluation Results and Analysis}

\begin{wraptable}{r}{0.3\linewidth}
    \vspace{-0.9em}
    \centering
    \small
    \caption{\textcolor{black}{Data allocation rule for PUR.}}
    \label{tab:lookup_peft_ratio}
    \begin{tabular}{cc}
        \toprule
        $S_{\mathrm{conc}}$ Range & PEFT Ratio \\
        \midrule
        0.66--0.70 & 30\% \\
        0.70--0.72 & 40\% \\
        0.72--0.74 & 50\% \\
        0.74--0.76 & 60\% \\
        0.76--0.80 & 70\% \\
        0.80--0.84 & 80\% \\
        \bottomrule
    \end{tabular}
    \vspace{-0.8em}
\end{wraptable}

\textcolor{black}{\noindent\textbf{Personalized LLM Generation Accuracy/Quality.} To study how the optimal allocation ratio changes across users with different semantic concentrations, we first conduct a probing study over a range of candidate allocation ratios. The results are shown in Table~\ref{tab:sconc_allocation}.
We make the following observations: (1) In every concentration interval, PUR outperforms all other methods when using the best allocation ratio for that interval, indicating that the workload split between PEFT and RAG is consistently beneficial. (2) The optimal allocation ratio varies across users with different semantic concentrations, suggesting that no single ratio is suitable for all users. (3) The optimal ratios do not span all candidate values, and extreme partition choices usually fail to fully realize the advantage of PUR. (4) Even when using ratios near the optimum, PUR often still outperforms all other methods, indicating that it is reasonably robust to the exact ratio choice. (5) The relationship between semantic concentration and the optimal allocation ratio is highly similar across the two tasks, suggesting that this allocation pattern has promising cross-task generalization potential. Based on these observations, we summarize a reference allocation rule for PUR, as shown in Table~\ref{tab:lookup_peft_ratio}, and use it as the default data partition rule in the subsequent experiments.}

\textcolor{black}{To better understand why PUR improves personalized generation accuracy/quality, we select users with different semantic concentrations and conduct a fine-grained analysis of their inference-time evaluation results, as shown in Table~\ref{tab:detailed_group_analysis_compact}. Based on PUR’s inference-time routing decisions, we partition each user’s test queries into PEFT-Region Queries (PRQ) and RAG-Region Queries (RRQ). First, compared with RAG, PUR performs significantly better on PRQ. A key reason is that PUR trains its personalized model on highly concentrated core information, which is consistent with our earlier observation that parametric learning can outperform RAG on highly concentrated data. On RRQ, PUR also relies on retrieval, but its responses are generated by a personalized model that has already learned the core information, allowing it to remain competitive with RAG. Second, compared with PEFT+RAG, PUR benefits from training its personalized model on less dispersed and noisy data. This cleaner training signal improves performance on both PRQ and RRQ, and ultimately leads to better overall performance.}

\textcolor{black}{\noindent\textbf{Personalized LLM Inference Latency.} Table~\ref{tab:detailed_group_analysis_compact} also reports the average inference latency for users at different semantic concentration levels. The results show that PUR not only achieves the best generation accuracy/quality, but also incurs lower inference latency than the retrieval-based methods. For User~1 with lower semantic concentration, PUR reduces the average inference time by 19.4\% and 14.4\% on the two tasks, respectively. For User~3 with higher semantic concentration, this latency advantage becomes even more pronounced, reaching 46.6\% and 42.5\%, respectively. This is because users with higher semantic concentration tend to have a larger fraction of queries routed to the PEFT region in PUR, allowing more responses to be generated without retrieval and thereby reducing average inference latency.}

\begin{table*}[t]
\centering
\caption{\textcolor{black}{Detailed LLM inference performance analysis on representative users with different semantic concentration levels. Each user's new queries are partitioned into PEFT-Region Queries (PRQ) and RAG-Region Queries (RRQ) according to PUR's inference path selection.}}
\label{tab:detailed_group_analysis_compact}
\resizebox{\textwidth}{!}{
\begin{tabular}{l|cccc|cccc|cccc}
\toprule
Metrics
& \makecell{PRQ \\ Perf.} 
& \makecell{RRQ \\ Perf.} 
& \makecell{Overall \\ Perf.}
& \makecell{Avg. Infer.\\Time (s) $\downarrow$}
& \makecell{PRQ \\ Perf.} 
& \makecell{RRQ \\ Perf.} 
& \makecell{Overall \\ Perf.}
& \makecell{Avg. Infer.\\Time (s) $\downarrow$}
& \makecell{PRQ \\ Perf.} 
& \makecell{RRQ \\ Perf.} 
& \makecell{Overall \\ Perf.}
& \makecell{Avg. Infer.\\Time (s) $\downarrow$}\\
\midrule
\multicolumn{13}{c}{\textbf{Personalized Movie Tagging (Acc. $\uparrow$)}} \\
\midrule
\multicolumn{1}{c|}{Method}
& \multicolumn{4}{c|}{User1 ($S_{\mathrm{conc}}$: 0.689 / PRQ Ratio: 34.5\%)}
& \multicolumn{4}{c|}{User2 ($S_{\mathrm{conc}}$: 0.752 / PRQ Ratio: 56.5\%)}
& \multicolumn{4}{c}{User3 ($S_{\mathrm{conc}}$: 0.811 / PRQ Ratio: 83.6\%)} \\
\midrule
PEFT      & 56.7 & 52.6 & 54.0 & \textbf{0.85} 
          & 57.7 & 55.0 & 56.5 & \textbf{0.90} 
          & \underline{68.3} & \underline{56.3} & 66.3 & \textbf{0.79} \\
RAG       & \underline{63.3} & \textbf{61.4} & \underline{62.1} & 1.75 
          & 61.5 & \underline{62.5} & 62.0 & 1.83 
          & 67.1 & \underline{56.3} & 66.3 & 1.78 \\
Full PEFT+RAG  & \underline{63.3} & \underline{57.9} & 59.8 & 1.77 
          & \underline{63.5} & \textbf{65.0} & \underline{64.1} & 1.81 
          & \underline{68.3} & \textbf{68.8} & \underline{68.4} & 1.74 \\
\textbf{PUR (Ours)}& \textbf{70.0} & \textbf{61.4} & \textbf{64.4} & \underline{1.41} 
          & \textbf{67.3} & \textbf{65.0} & \textbf{66.3} & \underline{1.29} 
          & \textbf{70.7} & \textbf{68.8} & \textbf{70.4} & \underline{0.93} \\

\midrule
\multicolumn{13}{c}{\textbf{Personalized Scholarly Title Generation (R-1 $\uparrow$)}} \\
\midrule
\multicolumn{1}{c|}{Method}
& \multicolumn{4}{c|}{User1 ( $S_{\mathrm{conc}}$: 0.701 / PRQ Ratio: 31.5\%)}
& \multicolumn{4}{c|}{User2 ( $S_{\mathrm{conc}}$: 0.743 / PRQ Ratio: 49.8\%)}
& \multicolumn{4}{c}{User3 ( $S_{\mathrm{conc}}$: 0.809 / PRQ Ratio: 81.1\%)} \\
\midrule
PEFT      & 44.1 & 47.7 & 46.5 & \textbf{1.25} 
          & 48.1 & 48.8 & 48.4 & \textbf{1.31} 
          & 49.8 & 49.0 & 49.6 & \textbf{1.29} \\
RAG       & \underline{47.5} & \underline{48.4} & \underline{48.1} & 2.57 
          & 49.1 & 48.9 & 49.0 & 2.64 
          & 49.4 & 49.3 & 49.4 & 2.51 \\
Full PEFT+RAG  & 46.3 & 47.4 & 47.1 & 2.64 
          & \underline{50.1} & \underline{50.3} & \underline{50.2} & 2.70 
          & \underline{50.3} & \underline{50.4} & \underline{50.3} & 2.66 \\
\textbf{PUR (Ours)}& \textbf{50.3} & \textbf{48.7} & \textbf{49.2} & \underline{2.20} 
          & \textbf{52.3} & \textbf{50.8} & \textbf{51.6} & \underline{1.93} 
          & \textbf{51.9} & \textbf{50.7} & \textbf{51.6} & \underline{1.53} \\
\bottomrule
\end{tabular}
}
\end{table*}

\begin{figure*}[t]
    \centering
    \begin{subfigure}[t]{0.23\textwidth}
        \centering
        \includegraphics[width=\linewidth]{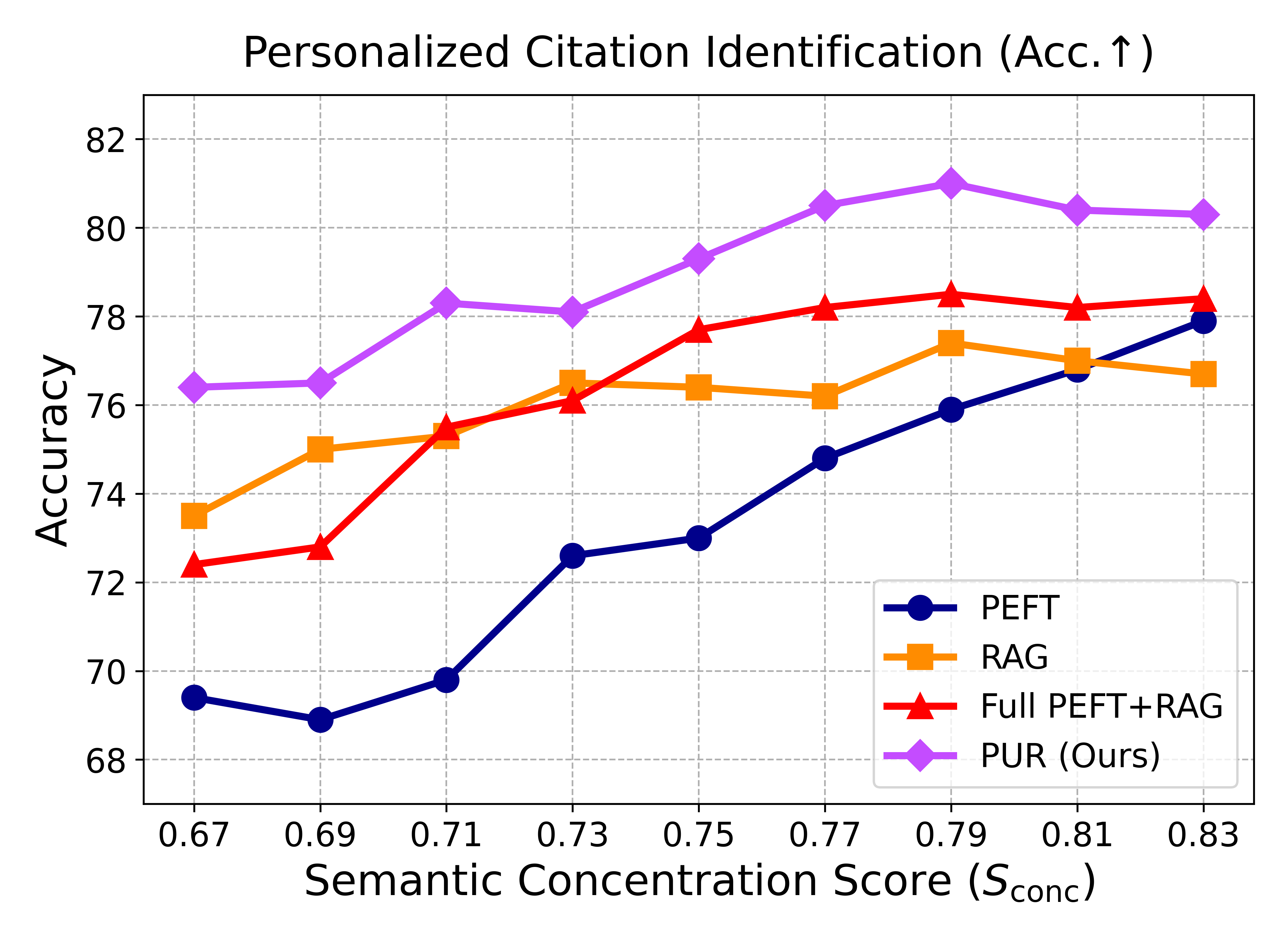}
        \label{fig:task1}
    \end{subfigure}
    \hfill
    \begin{subfigure}[t]{0.23\textwidth}
        \centering
        \includegraphics[width=\linewidth]{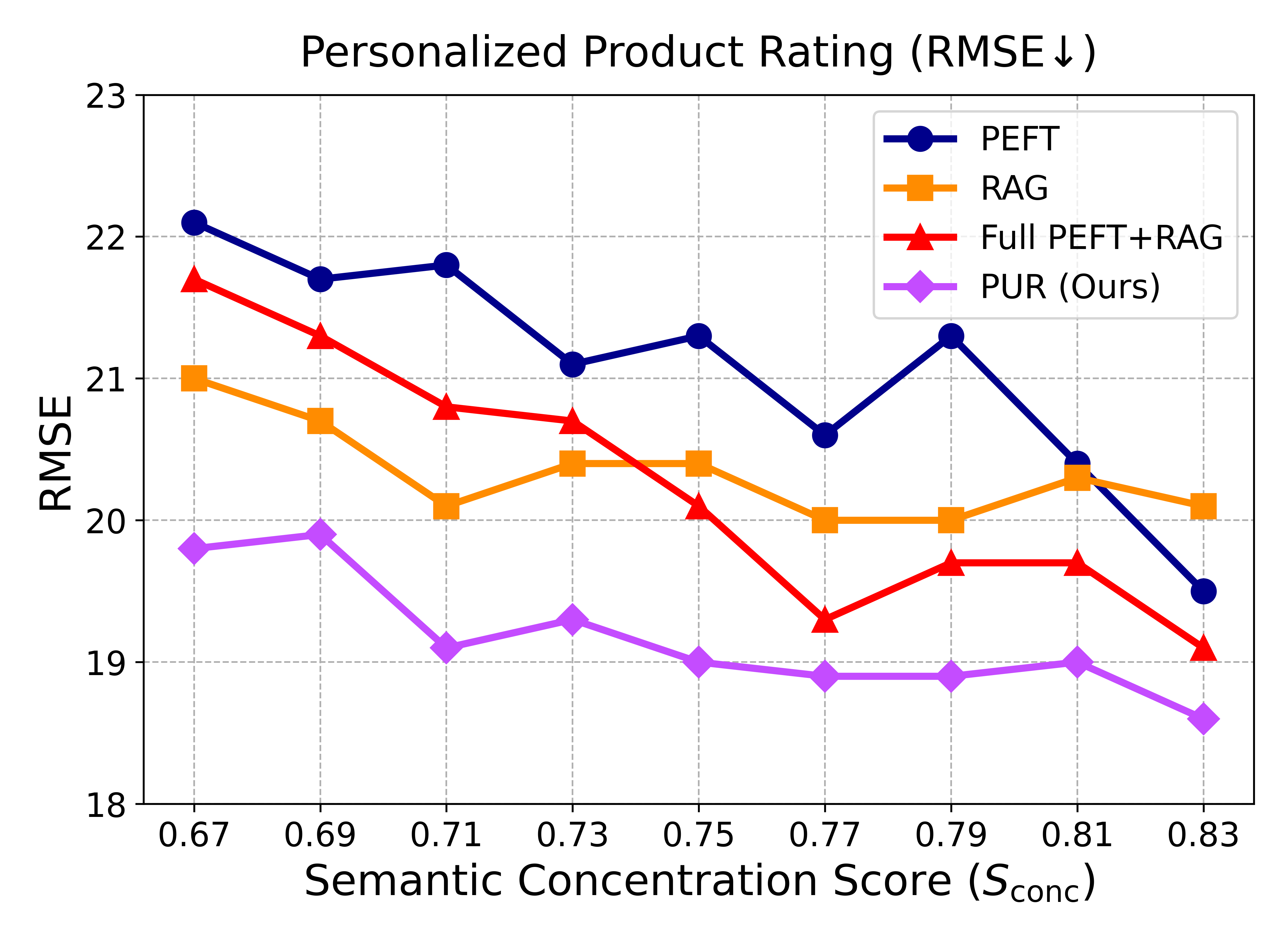}
        \label{fig:task2}
    \end{subfigure}
    \hfill
    \begin{subfigure}[t]{0.23\textwidth}
        \centering
        \includegraphics[width=\linewidth]{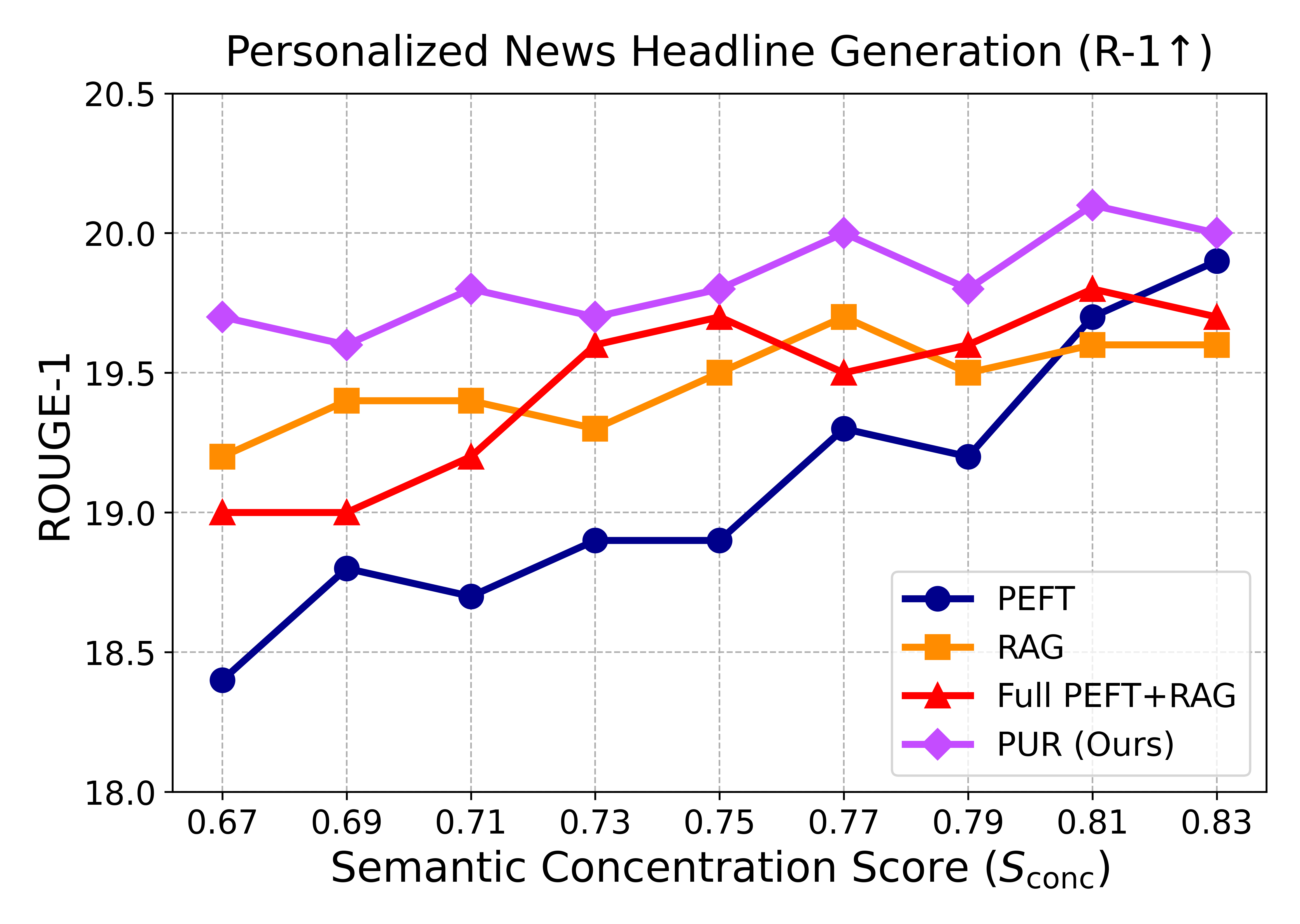}
        \label{fig:task2}
    \end{subfigure}
    \hfill
    \begin{subfigure}[t]{0.23\textwidth}
        \centering
        \includegraphics[width=\linewidth]{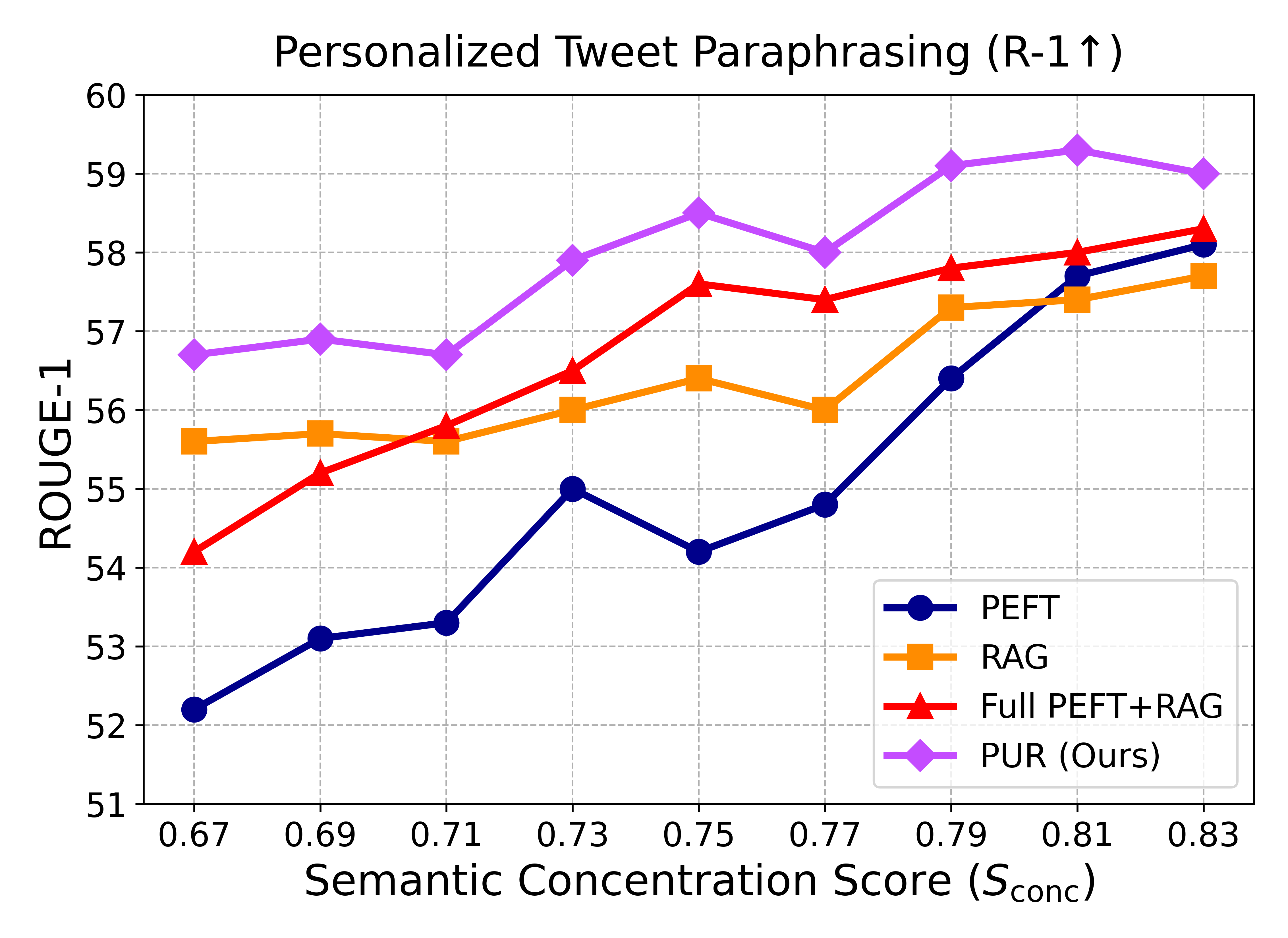}
        \label{fig:task2}
    \end{subfigure}
    \vspace{-0.8em}
    \caption{\textcolor{black}{Personalized generation accuracy/quality on the four remaining evaluated LaMP tasks not used in the probing study. PUR applies the allocation ratios obtained from the probing study to evaluate cross-task generalization.}}
    \label{fig:six_tasks}
    \vspace{-0.5em}
\end{figure*}

\textcolor{black}{\noindent\textbf{Generalization of the Workload Splitting Strategy.} To study how the allocation rule derived from the probing study generalizes beyond the probing setting, we further conduct experiments on the four remaining LaMP tasks that are not used in the probing study. As shown in Fig.~\ref{fig:six_tasks}, PUR consistently achieves the best performance across tasks and user groups with different semantic concentration levels, demonstrating strong cross-task and cross-user generalization. Although this empirical rule cannot guarantee that every user is assigned exactly their optimal allocation ratio, it still consistently outperforms other methods.}

\begin{figure*}[t]
    \centering
    \small
    \begin{subfigure}[t]{0.23\textwidth}
        \centering
        \includegraphics[width=\linewidth]{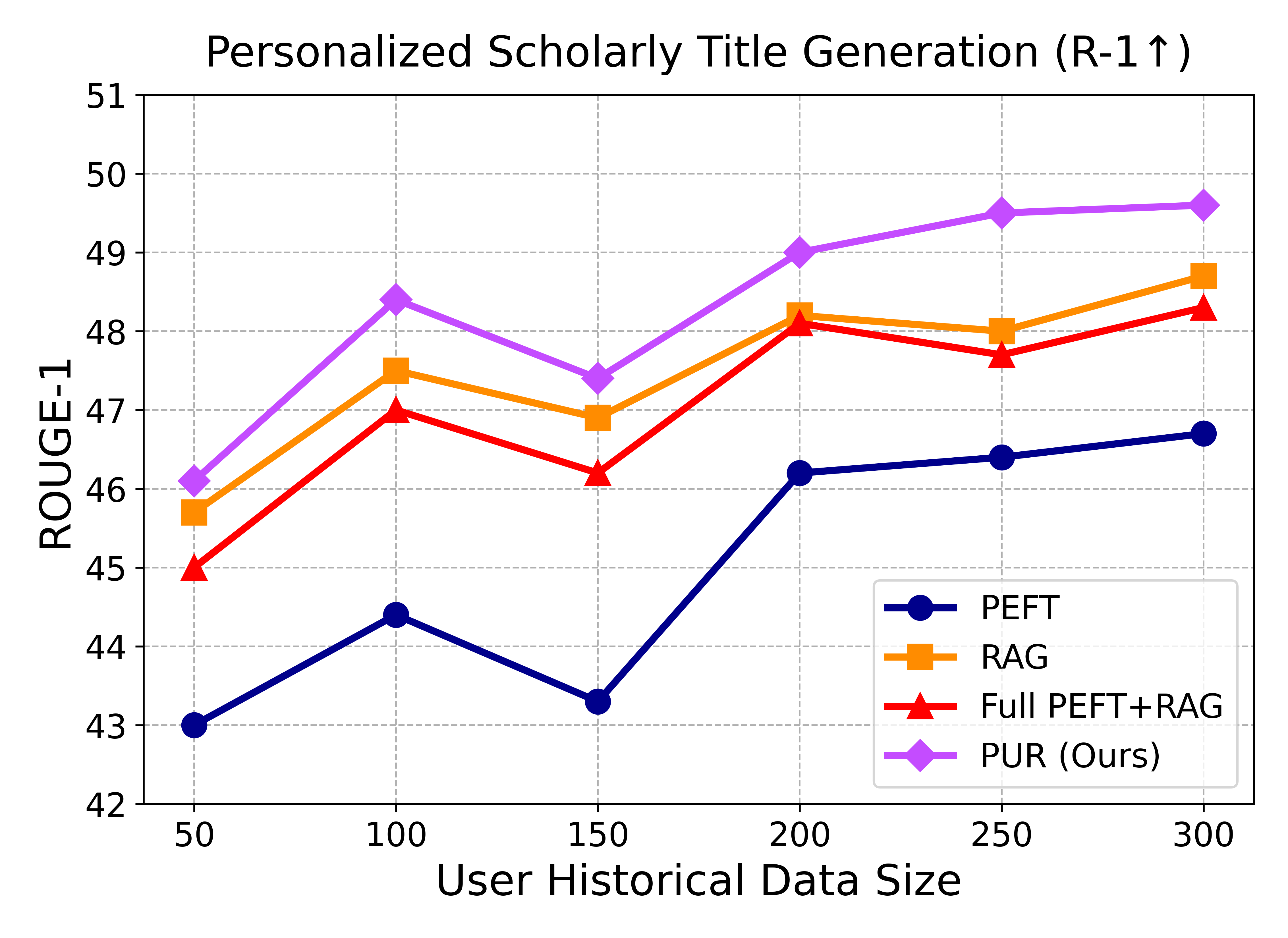}
        \caption{$S_{\mathrm{conc}}$: 0.68-0.70}
        \label{fig:datasize_task1}
    \end{subfigure}
    \hfill
    \begin{subfigure}[t]{0.23\textwidth}
        \centering
        \includegraphics[width=\linewidth]{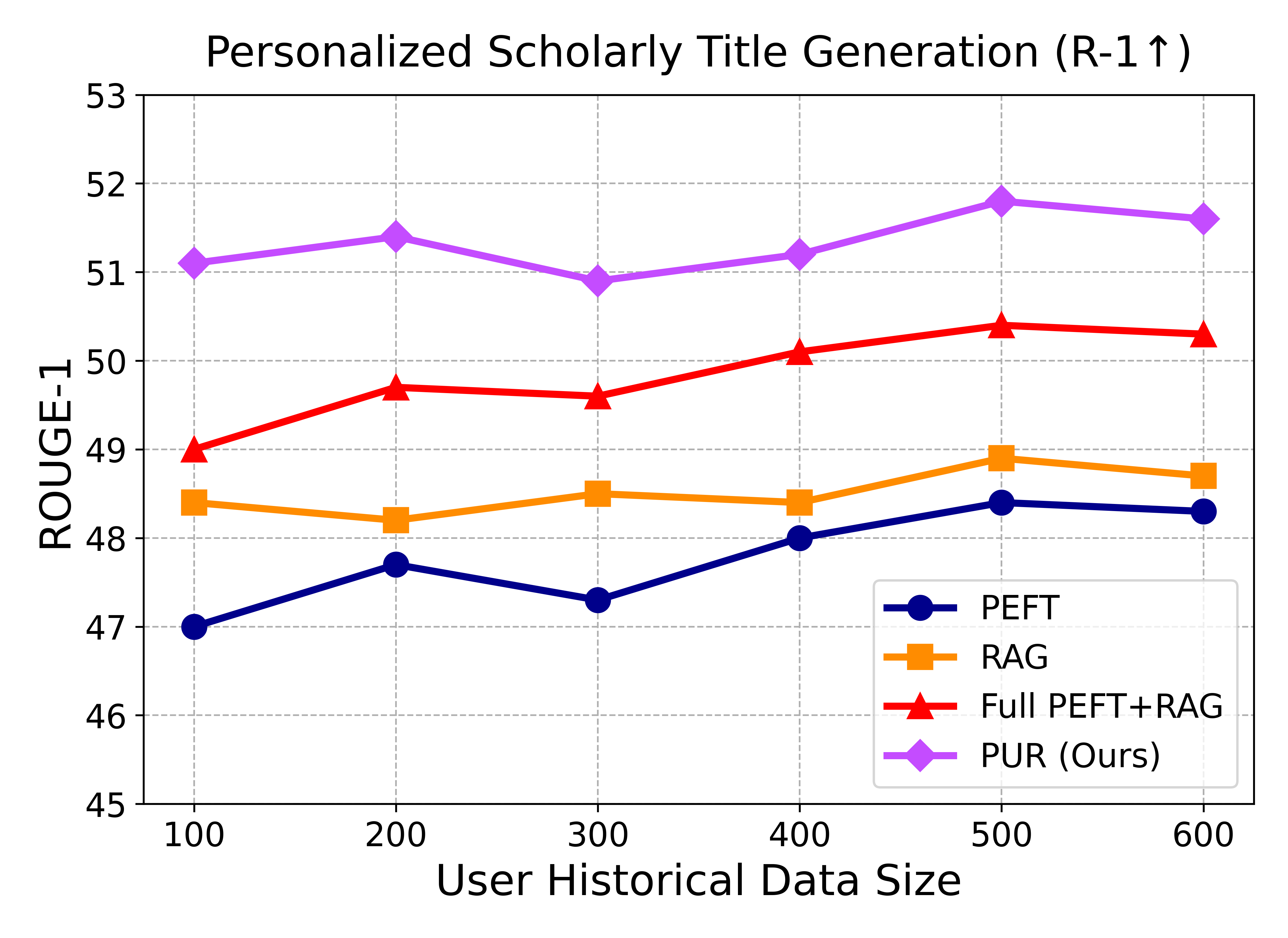}
        \caption{$S_{\mathrm{conc}}$: 0.74-0.76}
        \label{fig:datasize_task2}
    \end{subfigure}
    \hfill
    \begin{subfigure}[t]{0.23\textwidth}
        \centering
        \includegraphics[width=\linewidth]{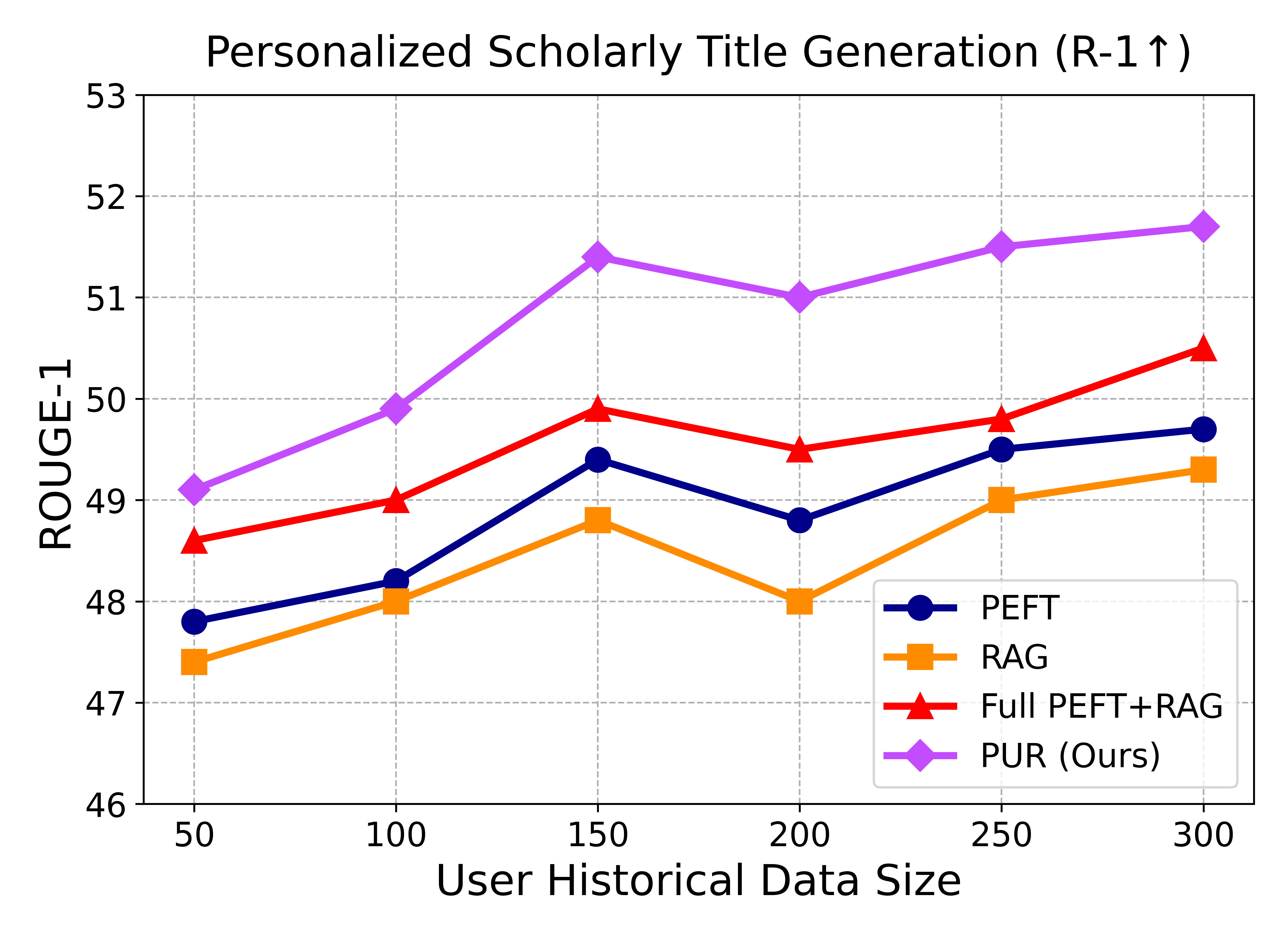}
        \caption{$S_{\mathrm{conc}}$: 0.80-0.82}
        \label{fig:datasize_task3}
    \end{subfigure}
    \hfill
    \begin{subfigure}[t]{0.23\textwidth}
        \centering
        \includegraphics[width=\linewidth]{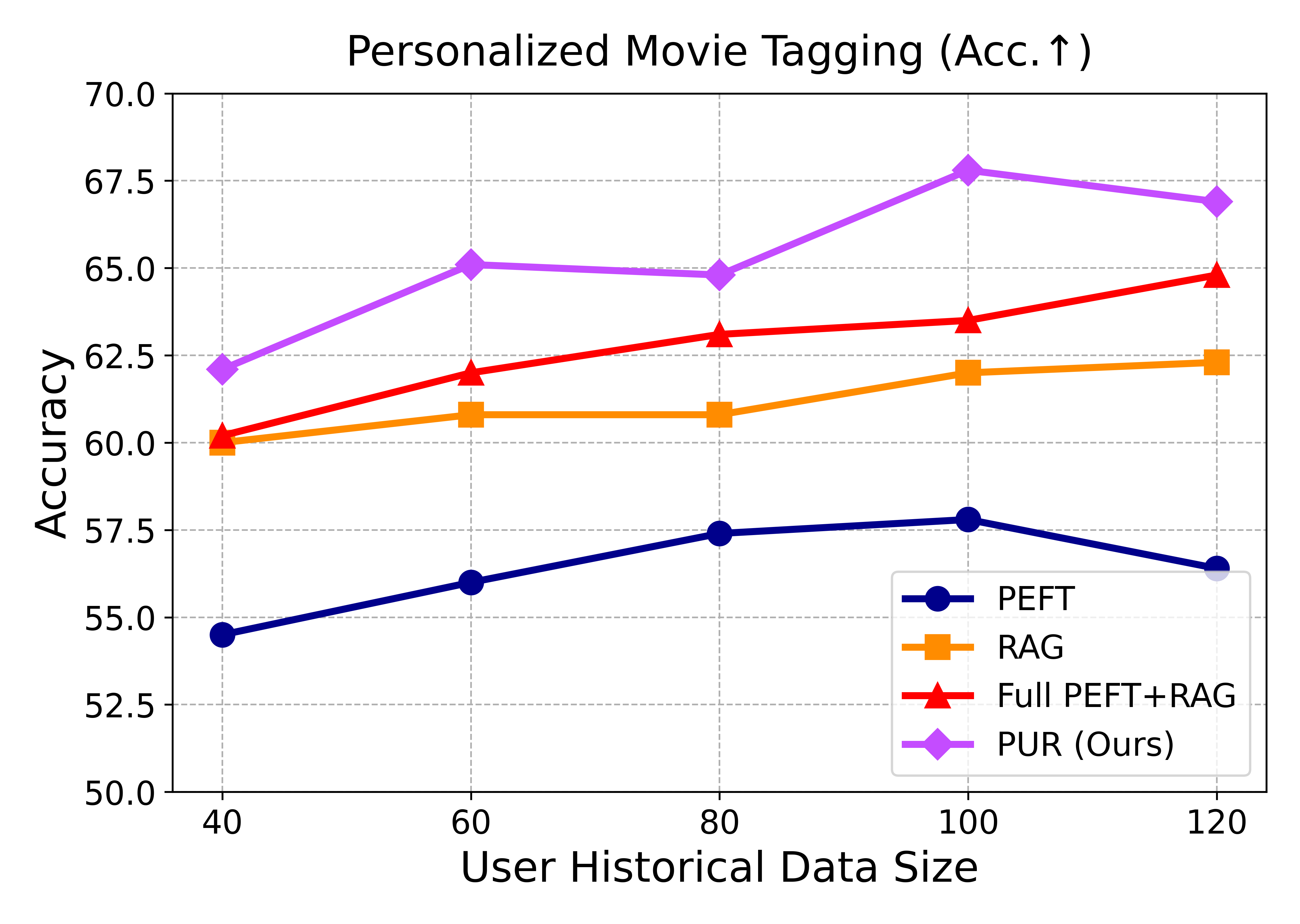}
        \caption{$S_{\mathrm{conc}}$: 0.74-0.76}
        \label{fig:datasize_task4}
    \end{subfigure}

    \caption{\textcolor{black}{Impact of historical data size on the relative performance of the methods. PUR consistently achieves the best performance across different user historical data sizes, while the relative ranking of the other methods remains largely unchanged within the same semantic concentration interval.}}
    \label{fig:history_number}
\end{figure*}

\textcolor{black}{\noindent\textbf{Impact of Historical Data Scale.} To study the impact of historical data size, we compare the performance of different methods across users with different amounts of historical data within the same semantic concentration interval. Figure~\ref{fig:history_number} shows the results across different tasks and concentration intervals. The results show that PUR consistently achieves the strongest performance across different historical data sizes. More importantly, unlike our earlier observations on semantic concentration, the relative ranking of the other methods remains largely unchanged as historical data size varies. This indicates that, for strategy selection, the distributional structure of historical data is a more important factor than its scale.}



\begin{figure*}[t]
    \centering

    \begin{subfigure}[t]{0.23\textwidth}
        \centering
        \includegraphics[width=\linewidth]{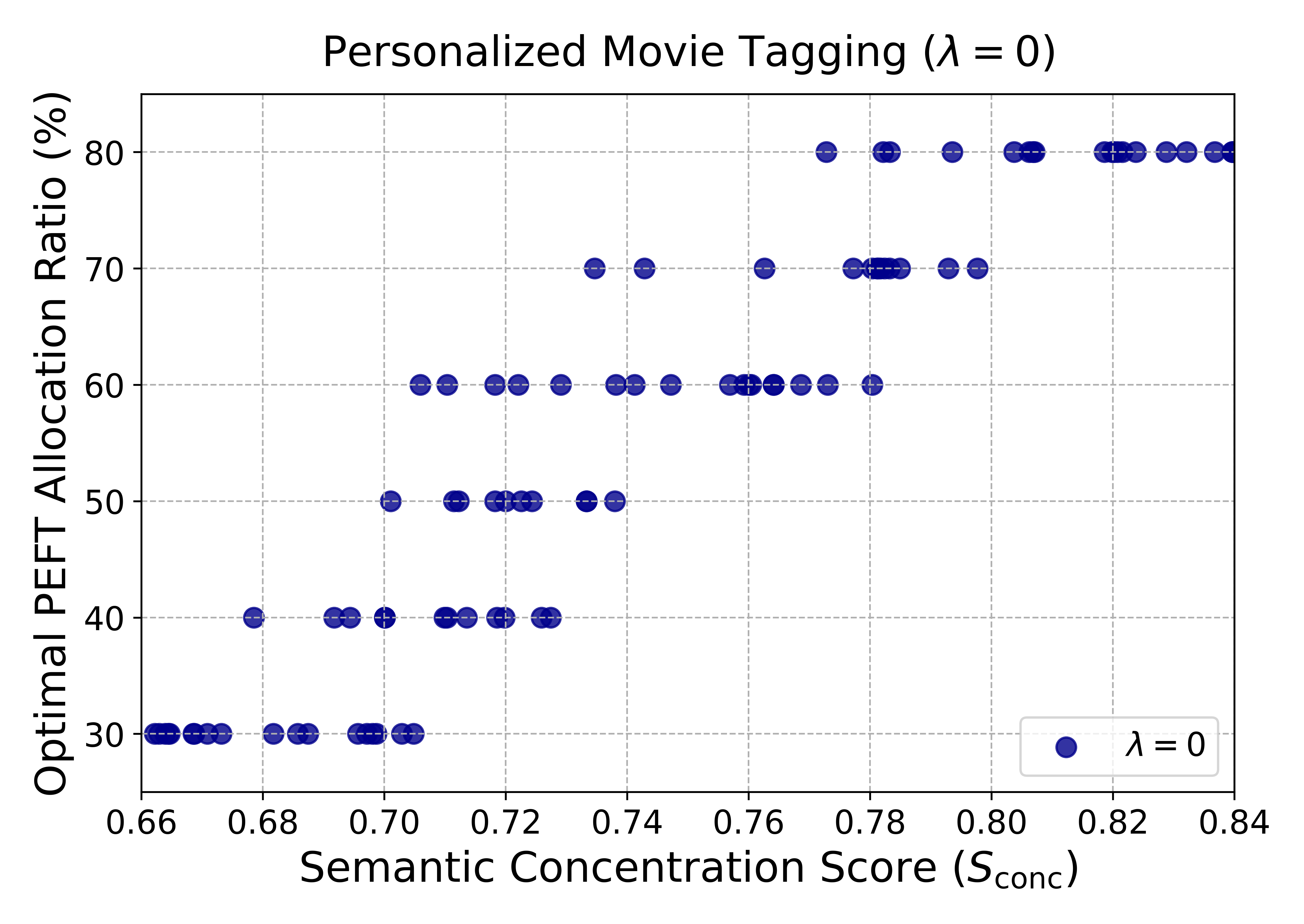}
        \label{fig:task1}
    \end{subfigure}
    \hfill
    \begin{subfigure}[t]{0.23\textwidth}
        \centering
        \includegraphics[width=\linewidth]{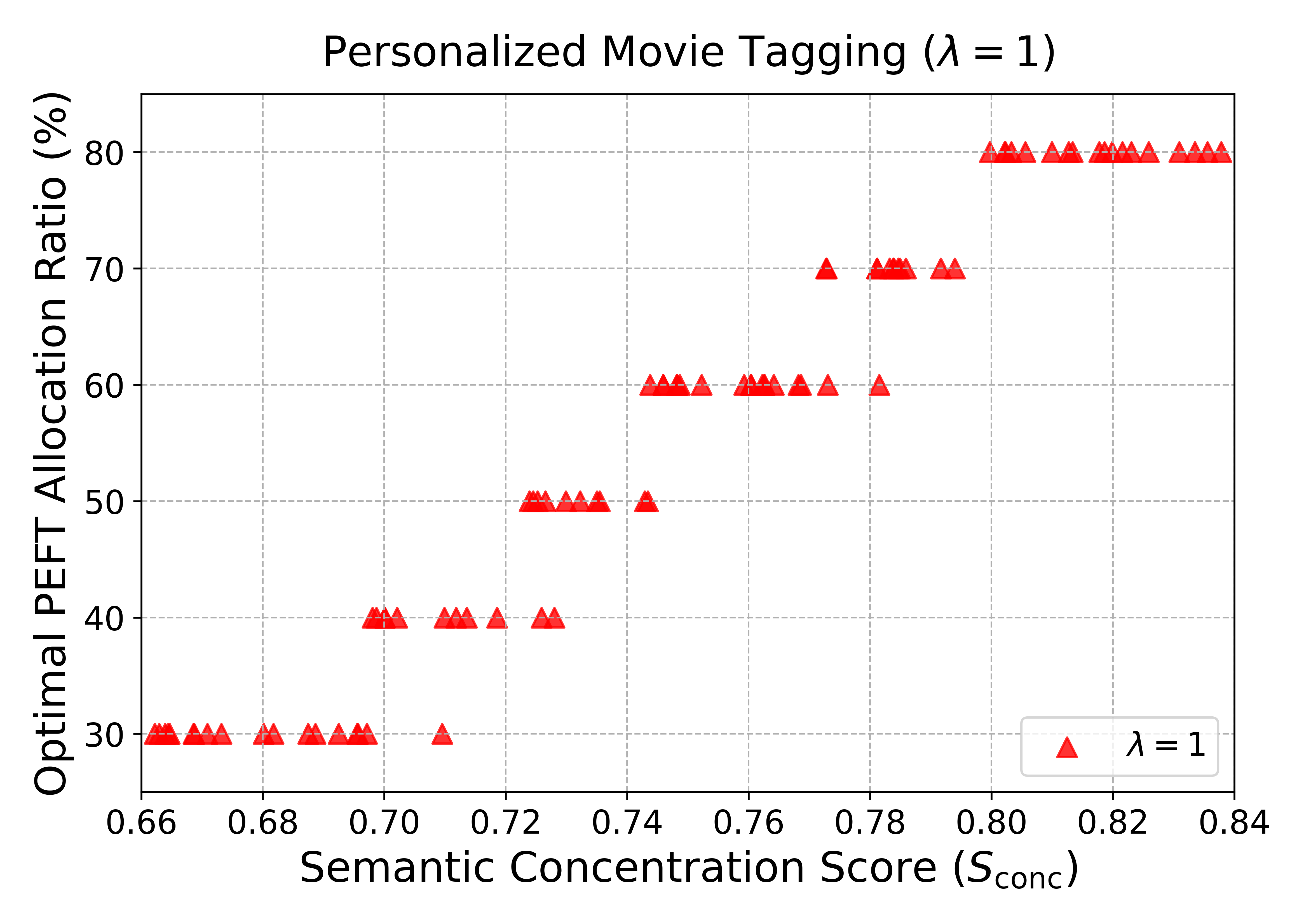}
        \label{fig:task2}
    \end{subfigure}
    \hfill
    \begin{subfigure}[t]{0.23\textwidth}
        \centering
        \includegraphics[width=\linewidth]{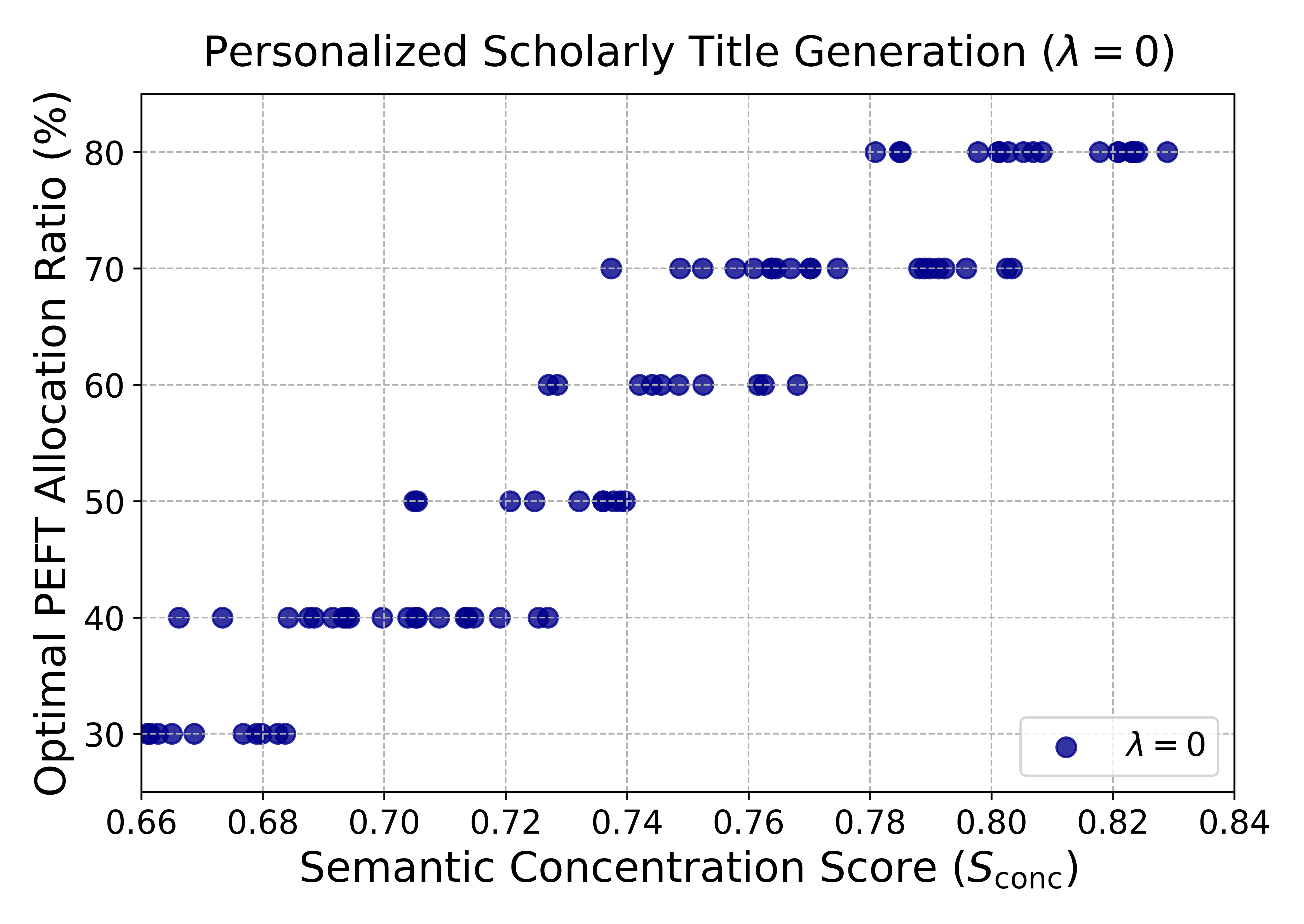}
        \label{fig:task2}
    \end{subfigure}
    \hfill
    \begin{subfigure}[t]{0.23\textwidth}
        \centering
        \includegraphics[width=\linewidth]{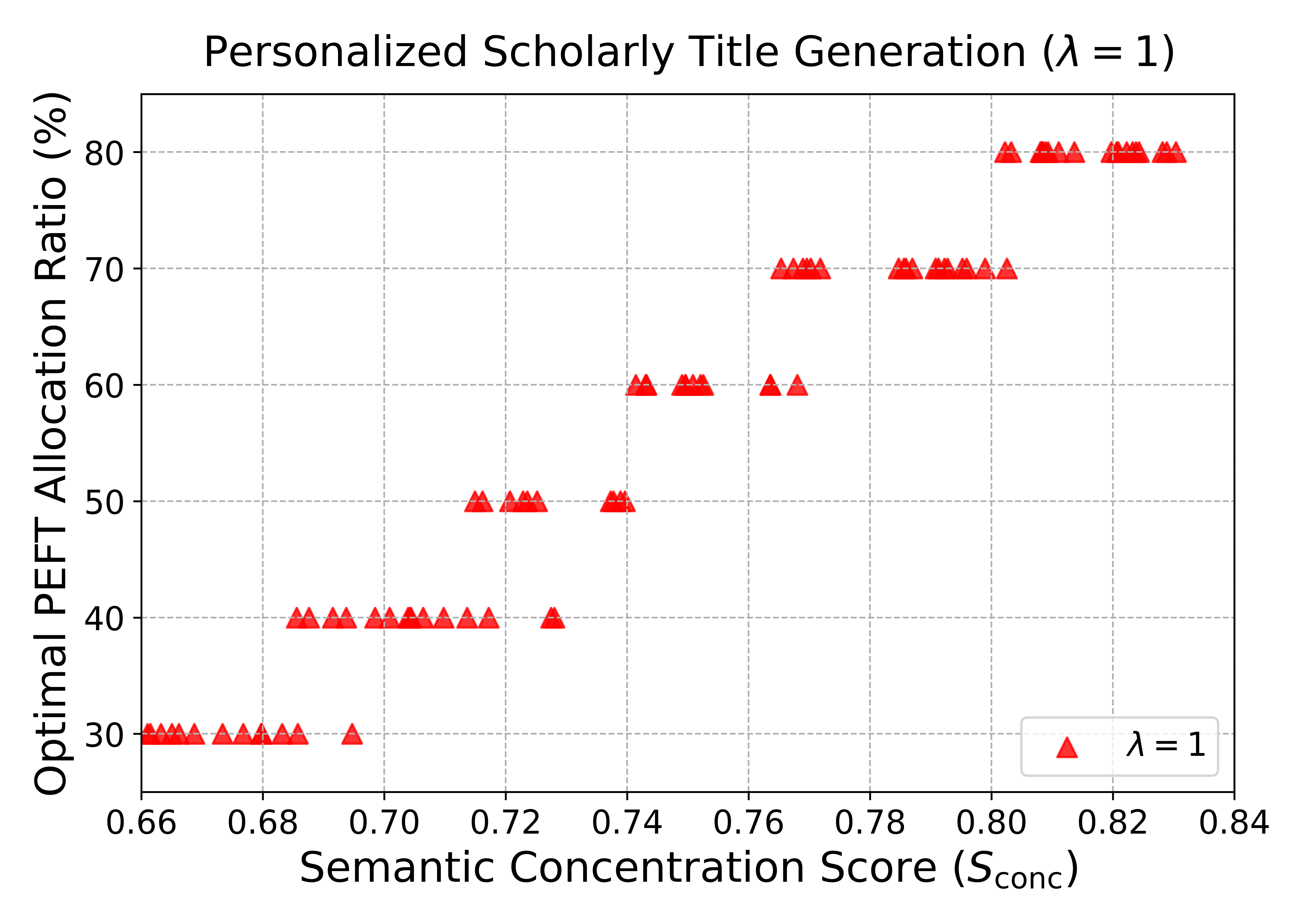}
        \label{fig:task3}
    \end{subfigure}

    \vspace{-1.1em}
    \caption{\textcolor{black}{Effect of the long-tail compensation term on the relationship between $S_{\mathrm{conc}}$ and the optimal PEFT allocation ratio. Each point represents a sampled user. Adding the compensation term ($\lambda=1$) yields better alignment in the score space than using average similarity alone ($\lambda=0$).}}
    \label{fig:lambda}
    \vspace{-1.0em}
\end{figure*}

\textcolor{black}{\noindent\textbf{Ablation Study of the Long-Tail Compensation Term.} To study the effect of the long-tail compensation term in $S_{\mathrm{conc}}$, we analyze how users’ $S_{\mathrm{conc}}$ values relate to their optimal PEFT allocation ratios under different values of $\lambda$. As shown in Fig.~\ref{fig:lambda}, when $\lambda=0$, i.e., without the long-tail compensation term, several users deviate from the overall $S_{\mathrm{conc}}$-to-allocation trend. A closer inspection (Appendix~\ref{sec:Further Analysis of the Long-Tail Compensation Term}) shows that these deviating users consistently have larger values of the tail-related statistic $\pi_{\mathrm{tail},i}\,h_{\mathrm{tail},i}$, indicating that average similarity alone is insufficient to provide a unified measure across users with different historical data distributions. When $\lambda=1$, the resulting scores become better aligned with the overall trend, and users sharing the same optimal allocation ratio become more tightly grouped in the score space. This indicates that the long-tail compensation term improves the reliability of $S_{\mathrm{conc}}$ as a signal for data allocation selection.}

\vspace{-0.4em}
\section{Conclusion}
\vspace{-0.4em}
\textcolor{black}{In this paper, we presented PEFT Union RAG (PUR), a novel ``fast and accurate'' personalized LLM approach that allocates user historical data between PEFT and RAG based on semantic concentration. Our empirical results show that such workload split between PEFT and RAG achieves better generation accuracy/quality than PEFT only, RAG only, or full PEFT+RAG, while reducing inference latency compared with the retrieval-based methods. The key reason is that PUR restricts PEFT training to concentrated core historical data, which reduces the impact of dispersed noisy information and leads to better generation quality. Meanwhile, queries whose required knowledge is covered by this core information can bypass retrieval, thereby reducing the averaged online LLM inference latency. More importantly, PUR takes the heterogeneity of users’ historical data distributions into account and adaptively determines the data allocation accordingly. Extensive experiments across multiple tasks show that PUR consistently outperforms other methods for users with different semantic concentration, while reducing inference latency by 14.4\%--46.6\% compared with the best performed compared method. These results collectively show that PUR is a practical and high-performing approach for leveraging historical data in personalized LLMs. }


\subsection*{AI Use Statement}

In this work, we did not use generative AI tools to formulate or refine research hypotheses, implement the research methods, generate or clean datasets, prove mathematical claims, or interpret experimental results. Tasks related to mathematical proofs and qualitative or thematic data analysis are not applicable to this work.

Additionally, we used generative AI tools to improve the readability of the manuscript and assist with formatting. All AI-assisted manuscript edits were reviewed and validated by the authors. We take full responsibility for the final content of this work, including the text, research claims, experimental design, data, experimental results, and related research artifacts.

\subsection*{Ethics Statement}

This work studies personalized LLM generation using users' historical data. All experiments are conducted on the existing LaMP personalization benchmark, and we did not recruit new human participants or directly collect any new personally identifiable or sensitive user information. The user historical data used in our experiments are obtained through the existing benchmark and are used solely for research and evaluation purposes. We do not attempt to identify, track, or re-identify individual users. 

Personalized LLM systems may raise privacy and data-governance concerns when deployed with real user histories. Our work focuses only on the allocation and inference strategies of PEFT and RAG under an existing research benchmark and does not involve the collection or deployment of real-world user data. Any practical deployment should follow applicable data licenses, privacy requirements, and user-consent mechanisms.

\bibliography{iclr2027_conference}
\bibliographystyle{iclr2027_conference}

\newpage

\appendix

\section{Further Experimental Results and Discussion}\label{sec:Appendix A}

\subsection{Applicability to Multi-Cluster Historical Data}\label{sec:Applicability to Multi-Cluster Historical Data}



\begin{table*}[!htbp]
\centering
\caption{\textcolor{black}{Performance comparison under multi-cluster historical data. Each team consists of two users whose merged historical data is clearly detected by HDBSCAN as forming two clusters. We report each user’s individual evaluation performance and the average performance over all queries from both users (Task: Personalized Scholarly Title Generation).}}
\label{tab:multi_cluster_team_results}
\resizebox{\textwidth}{!}{
\begin{tabular}{l|ccc|ccc|ccc}
\toprule
\multicolumn{1}{c|}{Team}
& \multicolumn{3}{c|}{($C_1$ $S_{\mathrm{conc}}$: 0.745, $C_2$ $S_{\mathrm{conc}}$: 0.752)}
& \multicolumn{3}{c|}{($C_1$ $S_{\mathrm{conc}}$: 0.811, $C_2$ $S_{\mathrm{conc}}$: 0.838)}
& \multicolumn{3}{c}{($C_1$ $S_{\mathrm{conc}}$: 0.769, $C_2$ $S_{\mathrm{conc}}$: 0.823)} \\
\cmidrule(lr){1-1} \cmidrule(lr){2-4} \cmidrule(lr){5-7} \cmidrule(lr){8-10}
\multicolumn{1}{c|}{Method}
& \makecell{User1 \\ Perf.}
& \makecell{User2 \\ Perf.}
& \makecell{Overall \\ Perf.}
& \makecell{User1 \\ Perf.}
& \makecell{User2 \\ Perf.}
& \makecell{Overall \\ Perf.}
& \makecell{User1 \\ Perf.}
& \makecell{User2 \\ Perf.}
& \makecell{Overall \\ Perf.} \\
\midrule
PEFT      & 47.5 & 48.3 & 47.9 & 49.1 & 49.9 & 49.5 & 48.0 & 49.5 & 48.9 \\
RAG       & 49.3 & 49.6 & 49.5 & 49.4 & 49.2 & 49.3 & 47.7 & 48.8 & 48.3 \\
Full PEFT+RAG  & \underline{50.1} & \underline{50.7} & \underline{50.4} 
               & \underline{50.8} & \underline{50.4} & \underline{50.6} 
               & \underline{48.6} & \underline{50.2} & \underline{49.5} \\
\textbf{PUR (Ours)} & \textbf{51.5} & \textbf{52.3} & \textbf{51.9} 
                    & \textbf{51.6} & \textbf{52.0} & \textbf{51.8} 
                    & \textbf{50.1} & \textbf{52.9} & \textbf{51.7} \\
\bottomrule
\end{tabular}
}
\end{table*}

\textcolor{black}{In the main experiments, a single user’s historical data typically exhibits a single-cluster structure under the current HDBSCAN setting. To further examine the applicability of PUR to multi-cluster data, we consider an additional personalized modeling scenario in which multiple team members share one personalized model. Since different members’ historical data may reflect different preferences or expression styles, the merged history may contain multiple relatively independent semantic centers.}

\textcolor{black}{To simulate this setting, we group users in pairs and run HDBSCAN on the merged historical data. We retain only those user pairs whose merged data are clearly detected as forming two clusters, and evaluate PUR against the other methods on these combinations. For each cluster, we apply the same concentration modeling and data allocation as in the main paper. The PEFT-selected samples from all clusters are then combined to fine-tune a single personalized model, rather than training separate PEFT models for individual clusters.}

\textcolor{black}{Table~\ref{tab:multi_cluster_team_results} reports the results for three user-pair combinations on the personalized scholarly title generation task. We report both the performance on each user’s individual evaluation queries and the average performance over all queries from the two users. The results show that PUR continues to outperform the other methods in these two-cluster settings, suggesting that the cluster-wise PUR design remains effective beyond the single-cluster case considered in the main experiments.}

\subsection{Effect of the Embedding Model}\label{sec:Effect of the Embedding Model}

\begin{table*}[!htbp]
\centering
\caption{\textcolor{black}{Performance comparison under different embedding models. For the same user groups, we report the resulting $S_{\mathrm{conc}}$ range, generation quality, and PUR’s PEFT subset overlap under different embedding models (Task: Personalized Scholarly Title Generation).}}
\label{tab:embedding_model_analysis}
\resizebox{\textwidth}{!}{
\begin{tabular}{l|l|ccc|ccc|ccc}
\toprule
\multicolumn{2}{c|}{}
& \multicolumn{3}{c|}{User Group 1}
& \multicolumn{3}{c|}{User Group 2}
& \multicolumn{3}{c}{User Group 3} \\
\cmidrule(lr){3-5} \cmidrule(lr){6-8} \cmidrule(lr){9-11}
\multicolumn{1}{c|}{Embedding model}
& \multicolumn{1}{c|}{Method}
& \makecell{$S_{\mathrm{conc}}$ \\ Range}
& \makecell{(R-1 $\uparrow$)}
& \makecell{PEFT Subset \\ Overlap}
& \makecell{$S_{\mathrm{conc}}$ \\ Range}
& \makecell{(R-1 $\uparrow$)}
& \makecell{PEFT Subset \\ Overlap}
& \makecell{$S_{\mathrm{conc}}$ \\ Range}
& \makecell{(R-1 $\uparrow$)}
& \makecell{PEFT Subset \\ Overlap} \\
\midrule

\multirow{4}{*}{BGE-Base-en-v1.5}
& PEFT      & -- & 46.4 & -- & -- & 48.3 & -- & -- & 49.6 & -- \\
& RAG       & -- & \underline{48.5} & -- & -- & 49.0 & -- & -- & 49.3 & -- \\
& Full PEFT+RAG  & -- & 48.3 & -- & -- & \underline{50.1} & -- & -- & \underline{50.3} & -- \\
& \textbf{PUR (Ours)} & 0.68-0.70 & \textbf{49.4} & 100\% & 0.74-0.76 & \textbf{51.5} & 100\% & 0.80-0.82 & \textbf{51.7} & 100\% \\

\midrule

\multirow{4}{*}{E5-Base-v2}
& PEFT      & -- & 46.4 & -- & -- & 48.3 & -- & -- & 49.6 & -- \\
& RAG       & -- & \underline{48.7} & -- & -- & 49.3 & -- & -- & 49.0 & -- \\
& Full PEFT+RAG  & -- & 48.3 & -- & -- & \underline{50.0} & -- & -- & \underline{49.9} & -- \\
& \textbf{PUR (Ours)} & 0.79-0.80 & \textbf{49.1} & 77.3\% & 0.81-0.82 & \textbf{51.8} & 83.6\% & 0.83-0.84 & \textbf{51.4} & 91.6\% \\

\midrule

\multirow{4}{*}{GTE-Base}
& PEFT      & -- & 46.4 & -- & -- & 48.3 & -- & -- & 49.6 & -- \\
& RAG       & -- & \underline{48.1} & -- & -- & 49.0 & -- & -- & 49.5 & -- \\
& Full PEFT+RAG  & -- & 48.0 & -- & -- & \underline{50.1} & -- & -- & \underline{50.1} & -- \\
& \textbf{PUR (Ours)} & 0.77-0.78 & \textbf{49.1} & 82.1\% & 0.80-0.82 & \textbf{51.6} & 89.5\% & 0.84-0.85 & \textbf{51.7} & 93.5\% \\

\bottomrule
\end{tabular}
}
\end{table*}

\textcolor{black}{Since PUR’s concentration modeling is built on the vector space induced by the embedding model and the corresponding similarity relations, we further evaluate PUR under different embedding models. In addition to the \texttt{bge-base-en-v1.5}~\citep{xiao2024c} model used in the main paper, we also consider \texttt{e5-base-v2}~\citep{wang2022text} and \texttt{gte-base}~\citep{li2023towards} models. Specifically, we select users from the personalized scholarly title generation task whose $S_{\mathrm{conc}}$ scores under the BGE embedding model fall into the intervals 0.68--0.70, 0.74--0.76, and 0.80--0.82. We then evaluate the same set of users under different embedding models.}

\textcolor{black}{First, as shown by the $S_{\mathrm{conc}}$ ranges for the same set of users in Table~\ref{tab:embedding_model_analysis}, the overall range of $S_{\mathrm{conc}}$ shifts across embedding models, but the relative relationships among users remain broadly consistent. This is expected, since different embedding models induce different representation spaces and similarity scales, making the absolute values of the concentration scores not directly comparable. What matters most is whether the scores preserve the relative concentration differences across users, rather than whether their numerical values are exactly the same.}

\textcolor{black}{Based on this observation, for the same set of users under the other two embedding models, we reuse the PEFT ratios derived under the BGE embedding model to partition the data, and then evaluate both generation quality and the overlap between the PEFT-selected subsets across embedding models. As shown in Table~\ref{tab:embedding_model_analysis}, PUR continues to outperform the competing methods for the same users under the other two embedding models, and the selected PEFT subsets also exhibit high overlap. These results suggest that the core mechanism of PUR remains stable across embedding models: data partitioning continues to exploit the complementary strengths of PEFT and RAG, and the conclusion that different users should adopt different allocation ratios remains unchanged. What may vary is the precise $S_{\mathrm{conc}}$-to-ratio mapping, which may require slight empirical recalibration under a new embedding model.}

\subsection{Computational Cost Analysis}
\begin{table*}[!htbp]
\small
\centering
\caption{\textcolor{black}{Computational cost comparison of different methods in the offline and online inference stages. We report the computational cost for the same three users analyzed in Table~\ref{tab:detailed_group_analysis_compact} on the Personalized Movie Tagging task.}}
\label{tab:comp_cost_analysis}
\resizebox{\textwidth}{!}{
\begin{tabular}{l|cc|cc|cc}
\toprule
\multicolumn{1}{c|}{}
& \multicolumn{2}{c|}{User 1 (Low $S_{\mathrm{conc}}$)}
& \multicolumn{2}{c|}{User 2 (Medium $S_{\mathrm{conc}}$)}
& \multicolumn{2}{c}{User 3 (High $S_{\mathrm{conc}}$)} \\
\cmidrule(lr){2-3} \cmidrule(lr){4-5} \cmidrule(lr){6-7}
\multicolumn{1}{c|}{Method}
& \makecell{Offline Comp. \\  (PFLOPs) $\downarrow$}
& \makecell{Inference Comp.\\ (TFLOPs) $\downarrow$}
& \makecell{Offline Comp. \\  (PFLOPs) $\downarrow$}
& \makecell{Inference Comp.\\ (TFLOPs) $\downarrow$}
& \makecell{Offline Comp. \\  (PFLOPs) $\downarrow$}
& \makecell{Inference Comp.\\ (TFLOPs) $\downarrow$} \\
\midrule
PEFT      & 1.096 & \textbf{30.7} & 1.107 & \textbf{34.7} & 1.209 & \textbf{36.1} \\
RAG       & \textbf{0.002} & 170.9 & \textbf{0.002} & 208.8 & \textbf{0.002} & 265.4 \\
Full PEFT+RAG  & 1.098 & 170.9 & 1.109 & 208.8 & 1.211 & 265.4 \\
\textbf{PUR (Ours)} & \underline{0.329} & \underline{122.5} & \underline{0.665} & \underline{110.4} & \underline{0.967} & \underline{73.7} \\
\bottomrule
\end{tabular}
}
\end{table*}

\textcolor{black}{We further compare the computational cost of different methods in both the offline stage and the online inference stage, measured in FLOPs. Table~\ref{tab:detailed_group_analysis_compact} reports the corresponding costs for the three users in Table~3 on the personalized movie tagging task.}

\textcolor{black}{The results show that PEFT and RAG exhibit clearly different computational cost profiles. For PEFT, the major cost is concentrated in the offline stage, since parameter training requires substantial computation. However, its online cost remains low since inference only uses a relatively short input context. RAG shows the opposite pattern. Its offline cost is small, but its online cost is much higher because retrieved texts make the input much longer, and the computation grows quickly with context length. Unlike the two extreme cost patterns above, PUR splits historical data between PEFT and RAG, making the computational cost more balanced across the offline and online stages.}

\begin{figure*}[t]
    \centering

    \begin{subfigure}[t]{0.473\textwidth}
        \centering
        \includegraphics[width=\linewidth]{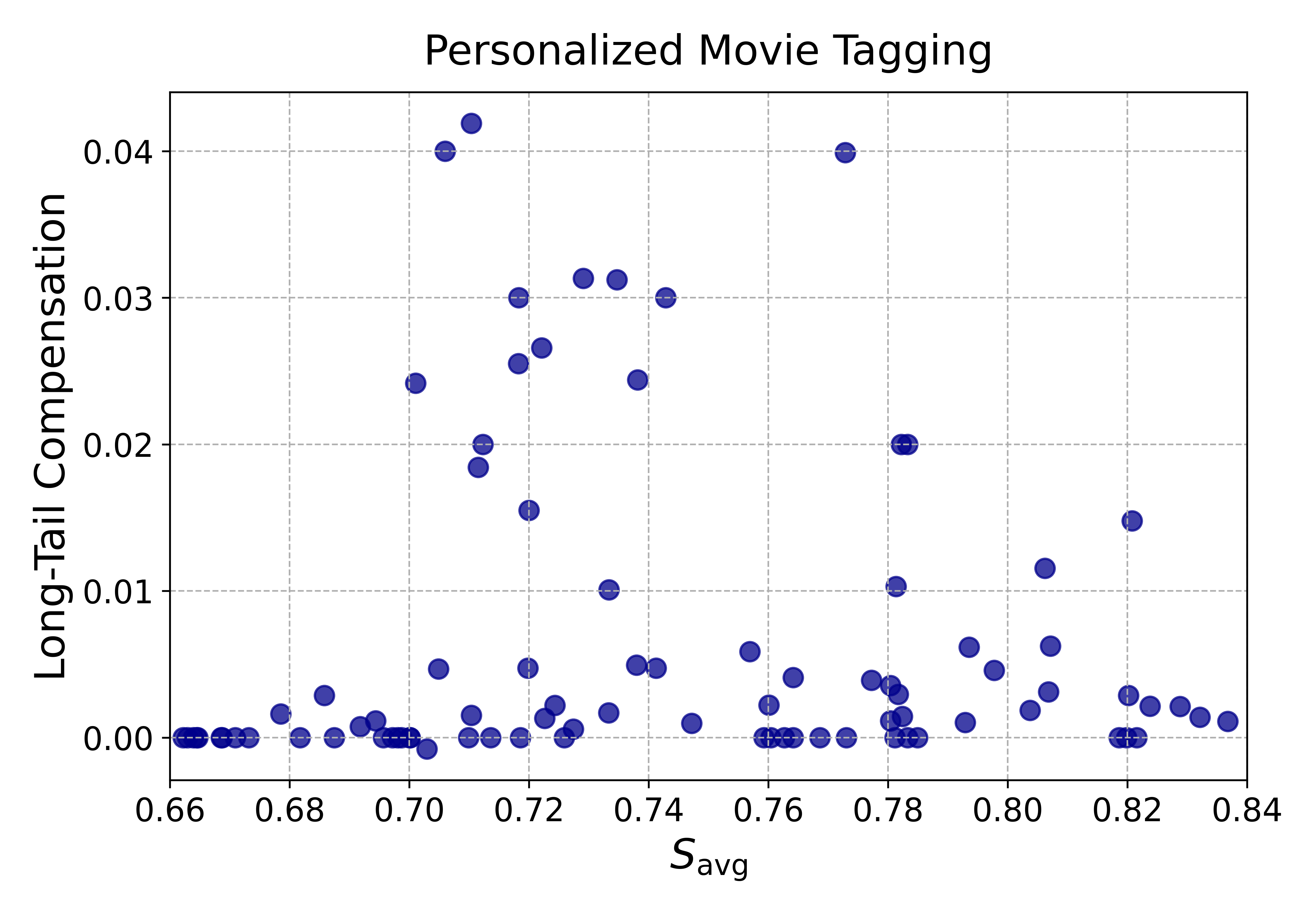}
        \label{fig:task1}
    \end{subfigure}
    \hfill
    \begin{subfigure}[t]{0.473\textwidth}
        \centering
        \includegraphics[width=\linewidth]{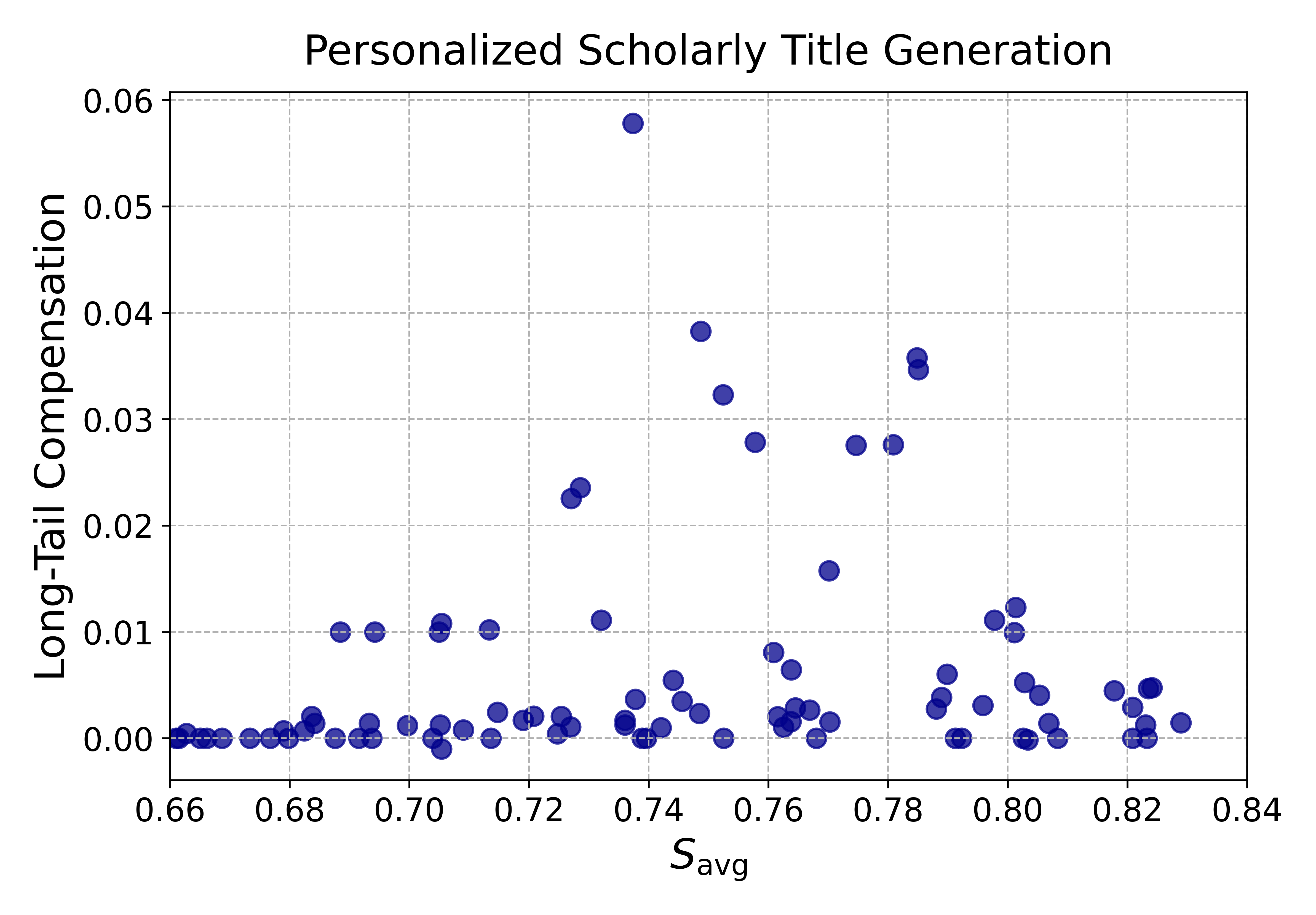}
        \label{fig:task3}
    \end{subfigure}

    \caption{\textcolor{black}{Distribution of $S_{\mathrm{avg}}$ and the long-tail compensation term $\pi_{\mathrm{tail},i}h_{\mathrm{tail},i}$ for the users involved in Fig.~\ref{fig:lambda}. Most users have relatively small compensation values, while a smaller subset exhibits substantially larger values and corresponds closely to the points that deviate from the overall trend when $\lambda=0$.}}
    \label{fig:Savg}
\end{figure*}

\subsection{Further Analysis of the Long-Tail Compensation Term}\label{sec:Further Analysis of the Long-Tail Compensation Term}

\textcolor{black}{We further analyze the distribution of the $S_{\mathrm{avg}}$ scores and the long-tail compensation term $\pi_{\mathrm{tail},i}h_{\mathrm{tail},i}$ for the users involved in Fig.~\ref{fig:lambda}. The results in Fig.~\ref{fig:Savg} show that, in real user data, most users have relatively small $\pi_{\mathrm{tail},i}h_{\mathrm{tail},i}$ values, while a smaller subset exhibits substantially larger values, indicating more pronounced long-tail structure in their historical data. These users with larger compensation terms closely align with the points that deviate from the overall trend in Fig.~\ref{fig:lambda} when $\lambda=0$. This suggests that, when a single $S_{\mathrm{conc}}$ score is used to characterize users with different data distribution patterns, relying only on $S_{\mathrm{avg}}$ tends to underestimate the appropriate PEFT allocation needs of users with pronounced long-tail structure. Overall, this further supports that the long-tail compensation term improves the reliability of $S_{\mathrm{conc}}$ as a signal for data allocation selection.}

\section{Dataset Details}\label{sec:Dataset Details and Hyperparameters}

\subsection{Historical Data Size}\label{sec:Historical Data Size}

\begin{table*}[!htbp]
\centering
\caption{\textcolor{black}{Average historical data size of the sampled evaluation users across different $S_{\mathrm{conc}}$ intervals and LaMP tasks.}}
\label{tab:avg_history_size_by_sconc}
\resizebox{0.9\textwidth}{!}{
\begin{tabular}{l|cccccc}
\toprule
\multicolumn{7}{c}{\textbf{Average historical data size of sampled users within each $S_{\mathrm{conc}}$ interval}} \\
\midrule
$S_{\mathrm{conc}}$ Range & LaMP-1 & LaMP-2 & LaMP-3 & LaMP-4 & LaMP-5 & LaMP-7 \\
\midrule
0.66--0.68 & 238 & 181 & 884 & 93 & 341 & 36 \\
0.68--0.70 & 335 & 126 & 1006 & 82 & 450 & 48 \\
0.70--0.72 & 403 & 124 & 946 & 119 & 580 & 59 \\
0.72--0.74 & 432 & 113 & 923 & 87 & 551 & 85 \\
0.74--0.76 & 388 & 137 & 891 & 127 & 528 & 77 \\
0.76--0.78 & 296 & 109 & 748 & 60 & 397 & 59 \\
0.78--0.80 & 289 & 98 & 410 & 91 & 331 & 99 \\
0.80--0.82 & 177 & 152 & 549 & 108 & 263 & 108 \\
0.82--0.84 & 210 & 152 & 326 & 151 & 225 & 151 \\
\bottomrule
\end{tabular}
}
\end{table*}

Here we present the average historical data size selected for each task in the main experiment across different S-score intervals to illustrate the scale of user data, as shown in Table~\ref{tab:avg_history_size_by_sconc}.

\subsection{Task Details}\label{sec:Task Details}

We present the task details as follows to help readers gain a better understanding of the task format.

\textbf{LaMP-1: Personalized Citation Identification} is a binary text classification task. Specifically, given user $u$ writes a paper $x$, the task aims to make the model determine which of the two candidate papers $u$ will cite in paper $x$ based on the user's history data, which contains the publications of user $u$.

\textbf{LaMP-2: Personalized Movie Tagging} is a 15-way text classification task to make tag assignments aligned with the user's history tagging preference. Specifically, given a movie description $x$, the model needs to predict one of the tags for the movie $x$ based on the user's historical movie-tag pairs.

\textbf{LaMP-3: Personalized Product Rating} is a 5-way text classification task and can also be understood as a regression task. Given the user $u$'s historical review and rating pairs and the input review $x$, the model needs to predict the rating corresponding to $x$ selected from 1 to 5 in integer. 

\textbf{LaMP-4: Personalized News Headline Generation} is a text generation task to test the model's ability to capture the stylistic patterns in personal data. Given a query $x$ that requests to generate a news headline for an article, as well as the user profile that contains the author's historical article-title pairs, the model is required to generate a news headline specifically for the given user.

\textbf{LaMP-5: Personalized Scholarly Title Generation} is a text generation task to test personalized text generation tasks in different domains. In this task, we require language models to generate titles for an input article $x$, given a user profile of historical article-title pairs for an author.

 \textbf{LaMP-7: Personalized Tweet Paraphrasing} is also a text generation task that tests the model's capabilities in capturing the stylistic patterns of authors. Given a user input text $x$ and the user profile of historical tweets, the model is required to paraphrase $x$ into $y$ that follows the given user's tweet pattern.

\end{document}